%% file: cvpr.tex
\documentclass[10pt,twocolumn,letterpaper,table]{article}
\usepackage{cvpr}
\usepackage{lmodern}
\usepackage{times}
\usepackage{enumitem}
\usepackage{graphicx}
\usepackage{amsmath}
\usepackage{amssymb}
\usepackage{tabularx}
\usepackage{multirow}
\usepackage{array}
\usepackage{tikz}
\usepackage{bm}
\usepackage{pgfplots}
\usetikzlibrary{shapes,arrows}
\usetikzlibrary{positioning}
\usetikzlibrary{decorations.pathreplacing}
\usetikzlibrary{external}
\usepackage[breaklinks=true,bookmarks=false]{hyperref}
\usepackage{cleveref}
\usepackage{booktabs}% http://ctan.org/pkg/booktabs
\usepackage{xparse}% http://ctan.org/pkg/xparse
\usepackage{nicefrac}

\pgfplotsset{compat=newest}
\usetikzlibrary{plotmarks}
\usepgfplotslibrary{patchplots}
\pgfplotsset{
	tick label style={font=\scriptsize},
	label style={font=\scriptsize},
	legend style={font=\scriptsize},
	title style={font=\scriptsize}}

\NewDocumentCommand{\rot}{O{90} O{0.5em} m}{\makebox[#2][c]{\rotatebox{#1}{#3}}}%

\newcommand{\ttb}[1]{{\scshape #1}\xspace}

\newcolumntype{x}{>\small c}
\newcommand{\bx}{\mathbf{x}}
\newcommand{\by}{\mathbf{y}}

\newcommand{\easy}{\textsc{Easy}\xspace}
\newcommand{\hard}{\textsc{Hard}\xspace}
\newcommand{\tough}{\textsc{Tough}\xspace}
\newcommand{\viewpoint}{\textsc{Viewpt}\xspace}
\newcommand{\illum}{\textsc{Illum}\xspace}
\newcommand{\sameseq}{\textsc{SameSeq}\xspace}
\newcommand{\diffseq}{\textsc{DiffSeq}\xspace}
\newcommand{\degrees}{\ensuremath{^{\circ}}\xspace}
\newcommand{\dataset}{HPatches\xspace}
\newlength\figH
\newlength\figW

\hypersetup{
  colorlinks,
  linkcolor={red!50!black},
  citecolor={blue!50!black},
  urlcolor={blue!80!black}
}

\newif\ifCLASSOPTIONcompsoc
\CLASSOPTIONcompsocfalse

\newif\ifarxiv
\arxivfalse

\newenvironment{customlegend}[1][]{%
  \begingroup
  \csname pgfplots@init@cleared@structures\endcsname
  \pgfplotsset{#1}%
}{%
  \csname pgfplots@createlegend\endcsname
  \endgroup
}%
\def\addlegendimage{\csname pgfplots@addlegendimage\endcsname}

\IfFileExists{/Users/vedaldi/.bashrc}{\cvprfinalcopy}{}

\cvprfinalcopy % *** Uncomment this line for the final submission
\def\cvprPaperID{2146} % *** Enter the CVPR Paper ID here
\ifcvprfinal\fi
\title{\dataset: A benchmark and evaluation of handcrafted and learned local descriptors}
\author{%
Vassileios Balntas$^{*}$\\
Imperial College London\\
{\tt\small v.balntas@imperial.ac.uk}
\and
Karel Lenc$^{*}$\\
University of Oxford\\
{\tt\small karel@robots.ox.ac.uk}
\and
Andrea Vedaldi\\
University of Oxford\\
{\tt\small vedaldi@robots.ox.ac.uk}
\and
Krystian Mikolajczyk\\
Imperial College London\\
{\tt\small k.mikolajczyk@imperial.ac.uk}\\
{\small $^{*}$Authors contributed equally to this work}
}

\begin{document}
% -----------------------------------------------------------

\maketitle
\begin{abstract}
%The field of local feature descriptors has seen remarkable growth in the recent years, leading to a significant amount of works that provide experimental evidence of improving over the state of the art, using a collection of the most common datasets and benchmarks in the field. Unfortunately, as we show, there is not a strong consistency in the evaluation results of most of the published work, due to the lack of a strict protocol for the experimental process. 
%In this paper, we introduce a novel dataset together with strictly defined protocols for evaluating feature descriptors. We evaluate the performance of several state of the art descriptors in three distinct tasks. This allows for comparisons that are more realistic and more representative of the different application scenarios for local feature descriptors.
In this paper, we propose a novel benchmark for evaluating local image descriptors. We demonstrate that the existing datasets and evaluation protocols do not specify unambiguously all aspects of evaluation, leading to ambiguities and inconsistencies in results reported in the literature. Furthermore, these datasets are nearly saturated due to the recent improvements in local descriptors obtained by learning them from large annotated datasets. Therefore, we introduce a new large dataset suitable for training and testing modern descriptors, together with strictly defined evaluation protocols in several tasks such as matching, retrieval and classification. This allows for more realistic, and thus more reliable comparisons in different application scenarios. We evaluate the performance of several state-of-the-art descriptors and analyse their properties. We show that a simple normalisation of traditional hand-crafted descriptors can boost their performance to the level of deep learning based descriptors within a realistic benchmarks evaluation.
\end{abstract}

\input{intro}
\input{datasets}
\input{hpatches}
\input{hbenchmark}
\input{experiments}
\input{conclusions}

%\clearpage
{\small
\bibliographystyle{ieee}
\bibliography{refs_clean}
%\bibliography{/Users/vedaldi/src/bibliography/bibliography}
}
\ifarxiv \input{appendix} \fi

%\clearpage\input{scratch}
% -----------------------------------------------------------
\end{document}

%% file: intro.tex
\ifCLASSOPTIONcompsoc\IEEEraisesectionheading{\section{Introduction}\label{sec:introduction}}\else\section{Introduction}\label{s:introduction}\fi

\ifCLASSOPTIONcompsoc\IEEEPARstart{L}{ocal }\else{Local }\fi feature
descriptors remain an essential component of image matching and
retrieval systems and it is an active area of research. With the
success of learnable representations and the availability of
increasingly-large labelled datasets, research on local descriptors
has seen a renaissance. End-to-end learning allows to thoroughly
optimise descriptors for available benchmarks, significantly
outperforming fully~\cite{lowe1999object} or semi-handcrafted
features~\cite{kmlinear, Trzcinski2013Binboost}.

\begin{table}[ht]
  \begin{center}
    \caption{Contradicting conclusions reported in literature while evaluating the {same descriptors} on the {same benchmark} (Oxford \cite{km2005pami}). Rows report inconsistent evaluation results due to variations of the implicit parameters e.g. of feature detectors.}\label{tab:inconsistencies}
    \footnotesize
    \vspace{0.5em}
    \setlength{\tabcolsep}{0.4em}
    \begin{tabular}{rl c rl}
      %\hline
      \ttb{LIOP} $>$ \ttb{SIFT} & \cite{miksik2012evaluation,wang2011local} &,& \ttb{SIFT} $>$ \ttb{LIOP} &  \cite{ASV}  \\
      \hline
      \ttb{BRISK} $>$ \ttb{SIFT} &  \cite{BRISK,miksik2012evaluation}  &,& \ttb{SIFT} $>$ \ttb{BRISK} &  \cite{LATCH}\\
      \hline
      \ttb{ORB}  $>$  \ttb{SIFT} &  \cite{rublee2011orb} &,& \ttb{SIFT} $>$ \ttb{ORB} &  \cite{miksik2012evaluation} \\
      \hline
      \ttb{BinBoost} $>$   \ttb{SIFT} &   \cite{LATCH,Trzcinski2013Binboost}  &,& \ttb{SIFT} $>$ \ttb{BinBoost} &  \cite{Balntas_2015_CVPR,ASV} \\
      \hline
      \ttb{ORB} $>$ \ttb{BRIEF} &  \cite{rublee2011orb} &,& \ttb{BRIEF} $>$ \ttb{ORB} &  \cite{LATCH} \\
      %\cline{1-2}
    \end{tabular}
  \end{center}\vspace{-2em}
\end{table}

Surprisingly however, the adoption of these purportedly better descriptors has been limited in applications, with SIFT~\cite{lowe1999object} still dominating the field. We believe that is due to the inconsistencies in reported performance evaluations based on the existing benchmarks~\cite{km2005pami, pickingthebestdaisy}. These datasets are either small, or lack diversity to generalise well to various applications of descriptors.
The progress in descriptor technology and application requirements has not been matched by a comparable development of benchmarks and evaluation protocols. As a result, while learned descriptors may be highly optimised for specific scenarios, it is unclear whether they work well in more general cases e.g.\ outside the specific dataset used to train them. In fact, just comparing descriptors based on published experiments is difficult and inconclusive as demonstrated in Table~\ref{tab:inconsistencies}.

In this paper, we introduce a novel benchmark suite for local feature 
descriptors, significantly larger, with clearly defined protocols and better 
generalisation properties, that can supersede the existing datasets. This is 
inspired by the success of the Oxford matching dataset~\cite{km2005pami}, the 
most  widely-adopted and still very popular benchmark for the evaluation of 
local features,  despite consisting of only 48 images. This is woefully 
insufficient for evaluating modern descriptors in the era of deep learning and 
large scale datasets. While some larger datasets exist, as discussed in 
\cref{s:related}, these have other important shortcomings in terms of data and 
task diversity, evaluation metrics and experimental reproducibility.
We address these shortcomings by identifying and satisfying crucial requirements from such a benchmark in \cref{s:design}.  

\textbf{Data diversity} is considered especially important for evaluating various properties of descriptors. To this end, we collect a large number of multi-image sequences of different scenes under real and varying capturing conditions, as discussed in \cref{s:bench-data}. Scenes are selected to be representative of different use cases and captured under varying viewpoint, illumination, or temporal conditions, including challenging nuisance factors often encountered in applications. The images are annotated with ground-truth transformations, that allow to identify unique correspondences necessary to assess the quality of matches established by descriptors.

\textbf{Reproducibility and fairness} of comparisons is crucial in benchmarks. This is addressed by eliminating the influence of detector parameters. Hence, the benchmark is based on extracted local image patches rather than whole images, which brings important benefits: i) it allows to compare descriptors modulus the choice of detectors, ii) it simplifies the process and makes the experiments reproducible, and, importantly, iii) it avoids various biases, e.g.\ the number or size of measurement regions or semi-local geometric constraints that make the results from image-based benchmarks incomparable (\cref{s:related}).

\textbf{Task diversity} is another requirement rarely addressed in exisiting evaluation benchmarks. To this end, we define three complementary benchmarking tasks in \cref{s:bench}: patch verification (classification of patch pairs), image matching, and patch retrieval. These are representative of different use cases and, as we show in the experiments, detectors rank differently depending on the task considered.

While this work is focused on local descriptors, the proposed dataset contains groundtruth, including pairwise geometric transformations, that will allow future evaluations of feature detectors as well. We believe that this benchmark will enable the community to gain new insights in state-of-the-art  local feature matching since it is more diverse and significantly larger than any existing dataset used in this field.
%The paper discusses in detail the existing benchmarks and their shortcomings (\cref{s:related}), the design of our benchmark (\cref{s:design}), the data collection (\cref{s:bench-data}) as well as the  tasks and protocols (\cref{s:bench-data}). 
We assess various methods including simple baselines, handcrafted ones, and state-of-the-art learned descriptors in \cref{s:experiments}. The experimental results show that descriptor performance and their ranking may vary in different tasks, and differs from the results reported in the literature. This further  highlights the importance of introducing a large, varied and reproducible evaluation benchmark for local descriptors.

All benchmark data and code implementing the evaluation protocols are made publicly available\footnote{\url{https://github.com/hpatches}}.

% In this paper, we first highlight the remarkable disagreement between several feature descriptor evaluations in terms of the state of the art across multiple previously published works, and we discuss some possible reasons for such significant differences. Subsequently, we briefly discuss the existing datasets and benchmarking methods in terms of their strengths and drawbacks. Lastly, we introduce a novel large-scale dataset suitable for both learning and evaluation local feature descriptors, that is the only dataset that exhibits all the desirable properties for benchmarking local feature descriptors. In addition, we describe three different clearly defined protocols for evaluating different aspects of feature descriptors using the newly introduced dataset. Lastly, we provide a performance review of several methods that were introduced in the recent years.

%% file: datasets.tex
\section{Review of existing benchmarks}\label{s:related}

In this section we review existing datasets and benchmarks for the evaluation of local  descriptors and discuss their main shortcomings.

\subsection{Image-based benchmarks}\label{s:image-match}

In image matching benchmarks, descriptors are used to establish correspondences between images of the same objects or scenes. Local features, extracted from each image by a co-variant detector, are matched by comparing their descriptors, typically with a nearest-neighbor approach. Then, putative matches are assessed for compatibility with the known geometric transformation between images (usually an homography) and the relative number of correspondences is used as the evaluation measure.

The most widely-adopted benchmark for evaluating descriptors and detectors is the~\emph{Oxford matching dataset}~\cite{km2005pami}. It consists of image sequences of 8 scenes, each containing 6 images, and ground-truth homographies. While the Oxford dataset contains images that are all captured by a camera, \emph{Generated Matching dataset}~\cite{Fischer2014} is obtained by generating images using synthetic transformations, and contains 16 sequences of 48 images. However, the synthetic nature of the transformations does not model all noise that typically occurs in the capturing process, thus making this data less challenging than the Oxford data~\cite{TFeat}. The \emph{DTU Robots dataset}~\cite{aanæsinteresting} contains real images of 3D objects, captured using a robotic arm in controlled laboratory conditions, which is suitable for certain application scenarios but of limited diversity in the data. The \emph{Hanover dataset}~{\cite{cordes2013high} investigates high-resolution matching and contains images of up to 8 megapixels with highly accurate ground-truth homographies. However, it is also limited by containing  only 5 scenes. The \emph{Edge Foci dataset}~\cite{zitnick2011edge} consists of sequences with very strong changes in viewing conditions, making the evaluation somewhat specialized to extreme cases; furthermore, the groundtruth for  non-planar scenes does not uniquely identify the correspondences since the transformations cannot be approximated well by homographies. Similarly, the \emph{WxBs dataset}~\cite{mishkin2015wxbs} focuses on very wide baseline matching, with extreme changes in geometry, illumination, and appearance over time.

All these datasets share an important shortcoming that leaves scope for variations in different descriptor evaluations: there is no pre-defined set of regions to match. As a consequence, results depend strongly on the choice of detector (method, implementation, and parameters), making the comparison of descriptors very difficult and unreliable. This is demonstrated in \cref{tab:inconsistencies} where different papers reach different conclusions even when they are evaluated on the same data using the same protocol.

Defining centre locations of features to match does not constrain the problem sufficiently either. For example, this does not fix the region of the image used to compute the descriptor, typically referred to as the \emph{measurement region}. Usually the measurement region is set to a fixed but arbitrarily set scaling of the feature size, and this parameter is often not reported or varies in papers. Unfortunately, this has a major impact on performance~\cite{simonyan2014learning}. Table~\ref{tab:sift_enlargment} shows matching scores for different scaling factors of the measurement region  in the Oxford data.\footnote{mAP is computed on the Leuven sequence in the Oxford matching dataset using the DoG detector and SIFT descriptor.} Variations of more than 50\% mAP occur; in fact, due to the planarity of such scenes, larger measurement regions lead to improved matching results. 

\begin{table}[t]
\begin{center}
\caption{Effect of using a different  $\rho$ to scale the size of the detected DoG keypoint to the size of the measurement region. 
% (mAP for the SIFT descriptor on the Leuven sequence int the Oxford matching benchmark; DoG detector).
%dataset. For this experiment we use the DoG detector. 
Columns 1\textbar X represent  matching scores between the first and the X image in the
sequence for different scaling factors $\rho$.}\label{tab:sift_enlargment}
{ \small
\begin{tabular}{lccccc}
\hline
$\rho$ & 1\textbar\ 2 & 1\textbar\ 3 & 1\textbar\ 4 & 1\textbar\ 5 &
1\textbar 6 \\
\hline
1   &  0.31  &  0.13  &  0.05  &  0.03  &  0.01 \\
%2   &  0.46  &  0.23  &  0.09  &  0.06  &  0.04 \\
4   &  0.68  &  0.44  &  0.24  &  0.15  &  0.11 \\
%8   &  0.74  &  0.57  &  0.43  &  0.32  &  0.24 \\
12  &  0.80  &  0.67  &  0.54  &  0.42  &  0.35 \\
20  &  0.87  &  0.77  &  0.69  &  0.55  &  0.50 \\
\hline
\end{tabular}
}
\end{center}\vspace{-2em}
\end{table}
In order to control for the size of the measurement region  and other important parameters such as the amount of blurring, resolution of the normalized patch used to compute a descriptor~\cite{vedaldi08vlfeat}, or use of semi-local geometric constraints, we argue that a descriptor benchmark should be based on \emph{image patches} rather than whole images. Thus, all such ambiguities are removed and a descriptor can be represented and evaluated as a function $f(\bm x) \in \mathbb{R}^{D}$ that maps a patch $\bm x \in \mathbb{R}^{H\times H\times 3}$ to a $D$-dimensional feature vector. This type of benchmark is discussed next.

\subsection{Patch-based benchmarks}\label{s:patch-matching}

Patch based benchmarks consist of patches extracted from interest point locations in images. The patches are then normalised to the same size, and annotated pair- or group-wise with labels that indicate positive or negative examples of correspondence. The annotation is typically established by using image groundtruth, such as geometric transformations between images. In case of image based evaluations the process of extracting, normalising and labelling patches leaves scope for variations and its  parameters differ between evaluations.

The first popular patch-based dataset was~\emph{PhotoTourism}~\cite{pickingthebestdaisy}. Since its introduction, the many benefits of using patches for benchmarking (\cref{s:image-matching}) became apparent. PhotoTourism introduced a simple and unambiguous evaluation protocol, which we refer to as \emph{patch verification}: given a pair of patches, the task is to predict whether they match or not, which reduces the matching task to a binary classification problem. This formulation is particularity suited for learning-based methods, including CNNs and metric learning in particular due to the large number patches available in this dataset. The main limitation of PhotoTourism is its scarce data diversity (there are only three scenes: Liberty, Notre-Dame and Yosemite), task diversity (there is only the patch verification task), and feature type diversity (only DoG features were extracted). The \emph{CVDS dataset}~\cite{CDVS} addresses the data diversity issue by extracting patches from five MPEG-CDVS: Graphics, Paintings, Video, Buildings and Common Objects. Despite its notable variety, experiments have shown that the state-of-the-art descriptors achieve high performance scores on this data~\cite{BalntasPhD}. The \emph{RomePatches dataset}~\cite{paulin2015local} consider a query ranking task that reflects image retrieval scenario, but is limited to $10K$ patches, which makes it an order of magnitude smaller than PhotoTourism.   

\subsection{Metrics}\label{s:metrics}

In addition to choosing data, patches, and tasks, the choice of evaluation metric is also important. For classification, the Receiver Operating Characteristic (ROC) curves have often been used~\cite{davis2006relationship,fawcett2004roc} as the basis for comparison. However, patch matching is intrinsically highly unbalanced, with many more negative than positive correspondence candidates; ROC curves are less representative for unbalanced data and, as a result, a strong performance in ROC space does not necessarily generalise to a strong performance in applications, such as the nearest-neighbor matching~\cite{simo2015deepdesc,ASV,Balntas_2015_CVPR,Trzcinski2012Dbrief}. Several papers~\cite{pickingthebestdaisy,Trzcinski2013Binboost,Trzcinski2012Dbrief} reported at a single point on the ROC curve (FPR95, i.e.\ the false positive rate at 95\% true positive recall) which is more appropriate for unbalanced data than the equal error rate or the area under the ROC curve; however, this reduces the information provided by the whole curve. Precision-Recall and mean Average Precision (mAP) are much better choices of metrics for unbalanced datasets -- for example \textsc{DBRIEF}~\cite{Trzcinski2012Dbrief} is excellent in ROC space but has very low  ($\approx 0.01$) mAP the Oxford dataset~\cite{LATCH}.

%%% Local Variables: 
%%% mode: latex
%%% TeX-master: "main"
%%% End: 

%% file: hpatches.tex
\section{Benchmark design}\label{s:design}
%We  discuss the requirements a benchmark should satisfy and the data.
\begin{table}[t]
  \begin{center}
    \caption{Comparison of existing datasets and the proposed \dataset dataset.\vspace{-0.5em}}\label{tab:our_dataset}
    {\small
\begin{tabular}{l cccccc }
  dataset & \scriptsize \rotatebox{40}{patch} & \scriptsize \rotatebox{40}{diverse} & \scriptsize \rotatebox{40}{real} &\scriptsize \rotatebox{40}{large} &\scriptsize \rotatebox{40}{multitask}  \\
  \hline
  Photo Tourism~\cite{LearningLocalImageDescriptors} & \checkmark &   & \checkmark & \checkmark &  \\
  DTU~\cite{aanæsinteresting} &  &    & \checkmark & \checkmark \\
  Oxford-Affine~\cite{km2005pami}&  & \checkmark &  \checkmark   \\
  Synth. Matching~\cite{Fischer2014}&  & \checkmark   & \checkmark \\
  CVDS~\cite{CDVS}& \checkmark & \checkmark  & & \checkmark & \\
  Edge Foci~\cite{zitnick2011edge} &   & \checkmark &   \checkmark & \\
  RomePatches~\cite{paulin2015local} & \checkmark & & \checkmark& \\
  RDED~\cite{CorRos2011a} &  & \checkmark &   \checkmark \\ \hline
  \textbf{\dataset} & \checkmark & \checkmark  & \checkmark & \checkmark & \checkmark \\
\end{tabular}
}
  \end{center}\vspace{-2.7em}
\end{table}

\begin{figure}[t]
\centering
\includegraphics[width=\columnwidth]{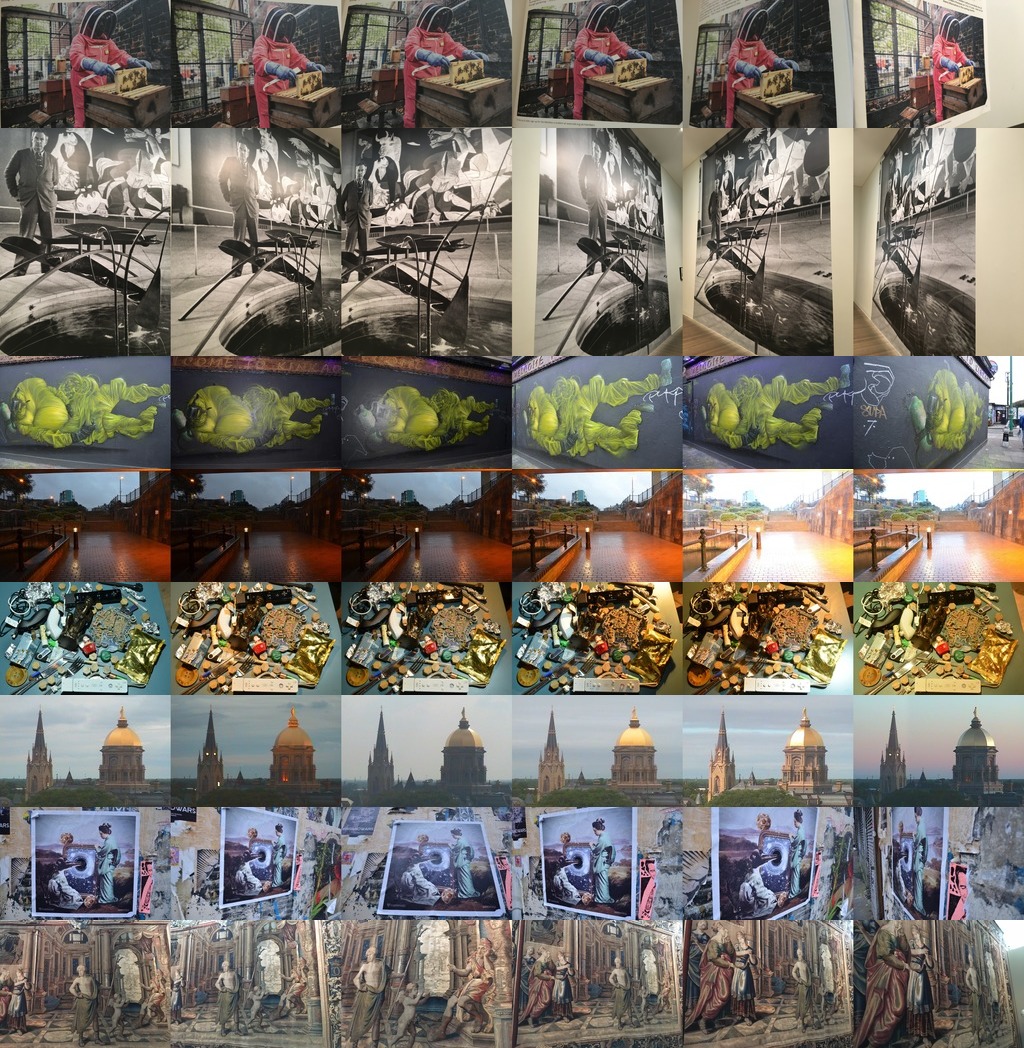}
\caption{Examples of image sequences; note the diversity of scenes and nuisance factors, including viewpoint, illumination, focus, reflections and other changes.\vspace{-1em}}
\label{fig:dataset}
\end{figure}

%\subsection{ Benchmark design goals}\label{s:design}
We address the shortcomings of the existing dataset, discussed in 
\cref{s:related}, by identifying the following requirements:
\begin{itemize}[noitemsep,topsep=2pt,parsep=0pt,partopsep=0pt,labelindent=0pt,itemindent=0pt,leftmargin=*]
\item \emph{Reproducible, patch-based}: descriptor evaluation should be done on patches to eliminate the detector related-factors. This leads to a standardisation across different works and makes results directly comparable.
\item \emph{Diverse}: representative of many different scenes and image capturing conditions.
\item \emph{Real}: real data have been found to be more challenging than a synthesized one due to nuisance factors that cannot be modelled in image transformations.
\item \emph{Large}: to allow accurate and stable evaluation, as well as to provide substantial training sets for learning based descriptors.
\item \emph{Multitask}: representative of several use cases, from matching image pairs to image retrieval. This allows cross-task comparison of descriptors performance within the same data.
%\item \emph{Patch-based}:  
\end{itemize}
Based on these desired properties, we introduce a new large-scale dataset of image sequences (\cref{s:bench-data}) annotated with homographies. This is used to generate a patch-based benchmark suite for evaluating local image descriptors (\cref{s:bench}). \Cref{tab:our_dataset} compares the proposed dataset to existing benchmarks in terms of the properties stated above. 

%\todo{say more?}
%\todo{Comparison of number of patches/images/scenes per dataset?}

%and highlight and shows that the proposed dataset is the only one that combines the desired properties of being large scale, unconstrained, diverse and leading to reproducible results. In particular, this dataset is 15 times larger than the Oxford benchmark \todo{What about all the other ones? Holydays? PhotoTurism? This needs to be written much more carefully}, the standard previously used equivalent real world matching dataset (Oxford matching), our dataset is 15 times larger, thus enabling a large set of potential novel uses, such as deep learning of local feature descriptors and detectors.

\section{Images and patches}\label{s:bench-data}

%\begin{figure}
%	\centering
%  \begin{tabular}{c c}
%    \rot{~~\easy} & \includegraphics[width=0.9\columnwidth]{figures/out/images_easy.eps} \\
%    \rot{~~\hard} &
%    \includegraphics[width=0.9\columnwidth]{figures/out/images_hard.eps} \\
%    \rot{~~\tough} &
%    \includegraphics[width=0.9\columnwidth]{figures/out/images_tough.eps}
%  \end{tabular}
%	\caption{Example of the selected geometry noise, visualized with the
%	 detected regions for the \easy, \hard and \tough distributions. Red ellipses are reprojected detections from the reference image.}
%	\label{fig:dets_easy_hard}
%\end{figure}

\begin{figure}[ht]
	\centering
  { \footnotesize
    \setlength{\tabcolsep}{0pt}
    \newcommand{\patchim}[1] {\includegraphics[trim={0 33em 0 0}, clip, width=0.072\columnwidth]{#1}}
\begin{tabular}{c  c  cccc  c  cccc  c  cccc}
  \patchim{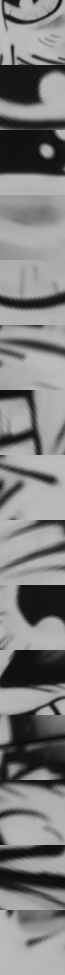} & &
  \patchim{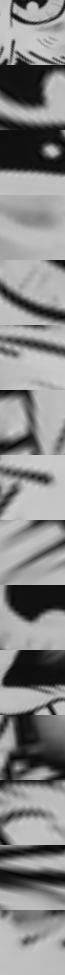} & 
  \patchim{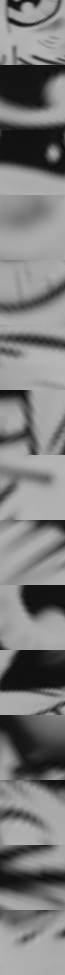} & 
  \patchim{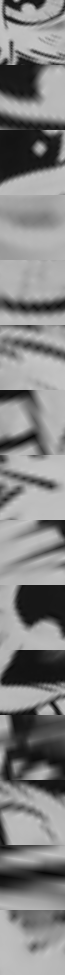} & 
  \patchim{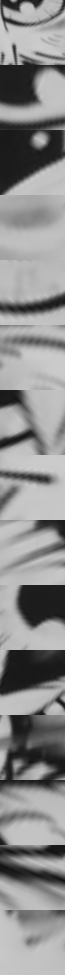} & 
  &
  \patchim{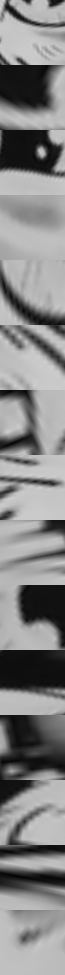} & 
  \patchim{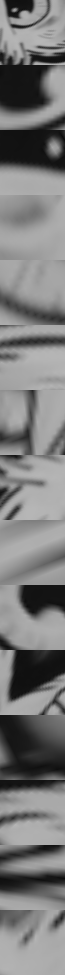} & 
  \patchim{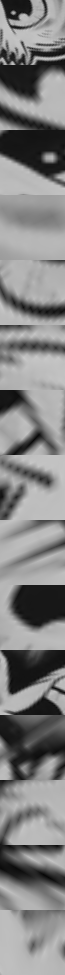} & 
  \patchim{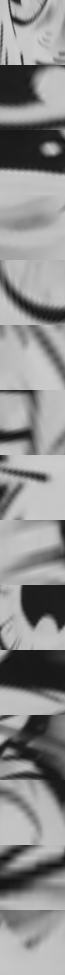} & 
  &
  \patchim{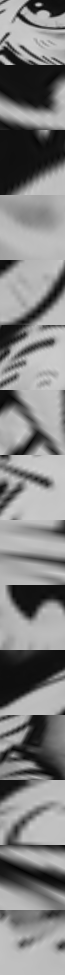} & 
  \patchim{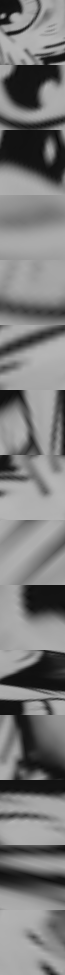} & 
  \patchim{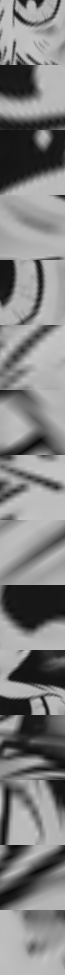} &
  \patchim{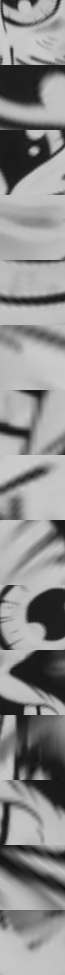} \\
    REF & ~~ & E1 & E2 & E3 & E4 & ~~ & H1 & H2 & H3 & H4 & ~~ & T1 & T2 & T3 & T4 \\
\end{tabular}
\vspace{0.5em}
}
	\caption{Example of the geometric noise visualized with the 
	extracted patches for the \easy, \hard and \tough distributions.\vspace{-1em}}
	\label{fig:patches_easy_hard}
\end{figure}

\paragraph{Images} are collected from various sources, including existing datasets. We have captured 51 sequences by a camera, 33 scenes are from \cite{jacobs2007consistent}, 12 scenes from \cite{aanæsinteresting}, 5 scenes from \cite{CorRos2011a}, 4 scenes from \cite{km2005pami}, 2 scenes from \cite{vonikakis2012improving} and 1 scene from \cite{asift}. Some of the sequences are illustrated in \cref{fig:dataset}. In 57 scenes the main nuisance factors are photometric changes and the remaining 59 sequences show significant geometric deformations due to viewpoint change.

A sequence %$\mathcal{D}=(L_0,L_1,\dots,L_K)$, 
includes a reference image and $5$ target images with varying photometric of geometric changes. The sequences are captured such that the geometric transformations between images can be well approximated by homographies from the reference image to each of the target images. The homographies are estimated following~\cite{km2005pami}. 

\paragraph{Patches} are extracted using the following protocol. Several scale invariant interest point detectors i.e.\ DoG, Hessian-Hessian and Harris-Laplace are used to extract features\footnote{VLFeat implementations of detectors are used.} for scales larger than $1.6px$, which give stable points. Near-duplicate regions are discarded based on their intersection-over-union (IoU) overlap ($> 0.5$) and one region per cluster is randomly retained. This keeps regions that overlap less than 0.5 IoU. Approximately 1,300 regions per image are then randomly selected.
% resulting in a total of $940 \times 10^3$ *****WHAT IS THIS NUMBER***** patches, which is approximately double the amount in PhotoTourism, the previously largest patch dataset.

%\begin{figure}
%	\centering
%	\setlength{\figH}{2cm}
%	\setlength{\figW}{0.9\columnwidth}
%	\input{figures/out/easy-hard-dist.tikz}
%	\caption{The distribution of the overlaps for the \easy and \hard geometry 
%	noise. The dashed line shows the mean overlap.}\label{fig:overlap}
%\end{figure}

\begin{figure}[t]
	\centering
	\setlength{\figH}{2.1cm}
	\setlength{\figW}{0.65\columnwidth}
	\input{figures/out/det-overlaps-adjusted.tex}
	\caption{The average overlap accuracy of Hessian and Hessian-Affine detector on the 
	viewpoint sequences of the \cite{Mikolajczyk05}. Line color encodes  
	dataset and line style a detector.  The selected 
	overlaps of the \easy and \hard variants are visualised with a dotted line.\vspace{-1em}}\label{fig:det-overlap}
\end{figure}
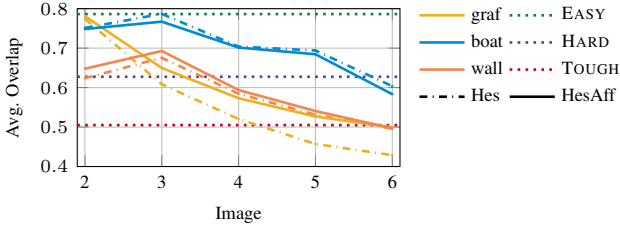

For each sequence, patches are detected in the reference image  and projected on the target images using the ground-truth homographies. This sidesteps the limitations of the detectors, which may fail to provide corresponding regions in every target images due to significant viewpoint or illumination variations.  Furthermore, it allows to extract more patches thus better evaluate descriptors in such scenarios. Regions that are not fully contained in all target images are discarded. Hence, a set of corresponding patches contains one from each image in the sequence.
In practice, when a detector extracts corresponding regions in different images, it does so with a certain amount of noise. In order to simulate this noise, detections are perturbed using three settings: \easy, \hard and \tough. This is obtained by applying a random transformation $T : \mathbb{R}^2 \to \mathbb{R}^2$ to the region before projection. Assuming that the region centre is the coordinate origin, the transformation includes rotation $R({\theta})$ by angle $\theta$, anisotropic scaling by $s / \sqrt{a}$ and $s\sqrt{a}$, and translation by $[m~t_x,m~t_y]$,
%\begin{equation} \label{eq:tf}
%T = 
%	R({\theta}) \cdot
%	\begin{bmatrix}
%		s / \sqrt{a} & 0 & m~t_x \\
%		0 & s \cdot \sqrt{a} & m~t_y
%	\end{bmatrix},
%\end{equation}
%where $R({\theta})$ is a rotation of angle $\theta$ and 
thus the translation is proportional to the detection scale $m$. The transformation parameters are uniformly sampled from the intervals
$\theta \in \left[ -\theta_{max}, \theta_{max} \right]$,
$t_x, t_y \in \left[ -t_{max}, t_{max} \right]$,
$\log_2(s) \in \left[ {-s_{max}}, {s_{max}} \right]$,
$\log_2(a) \in \left[ {-a_{max}}, {a_{max}} \right]$,
whose values for each setting are given in \cref{tab:gtf_params}.
% Examples of detected features and their perturbations three settings are given in~\cref{fig:overlap}. The difficulty of each band can be gauged by looking at the distribution of feature overlaps \cref{fig:dets_easy_hard}. 
These settings reflect the typical overlap accuracy of the Hessian and Hessian-Affine detectors on Oxford matching benchmark. There, images in each sequence are sorted by increasing transformation, resulting in increased detector noise. \Cref{fig:det-overlap} shows that the \easy, \hard, and \tough groups correspond to regions extracted in images 1-2, 3-4 and 5-6 of such sequences.

\begin{table}[ht]
	\caption{Range of geometric noise distributions, in units of a patch 
	scale.}\label{tab:gtf_params}
	\centering
	\small
	\begin{tabular}{l | c | c | c | c  } 
		Variant & $\theta_{max}$ & $t_{max}$ & $s_{max}$ & $a_{max}$ \\ \hline
		\easy   & $10 \degrees$  & 0.15      & 0.15      & 0.2 \\
		\hard   & $20 \degrees$  & 0.3       & 0.3       & 0.4 \\
		\tough  & $30 \degrees$  & 0.45      & 0.5       & 0.45 \\
	\end{tabular}
\end{table}

Detected regions are scaled with a factor of 5 (see \cref{s:related}). The smallest patch size in the reference image is $16 \times 16$px since only regions from detection scales above 1.6px are considered. In each region the dominant orientation angle is estimated using a histogram of gradient orientations \cite{lowe1999object}. Regions are rectified by normalizing the detected affine region to a circle using bilinear interpolation and extracting a square of $65 \times 65$ pixels. Example of extracted patches are shown in  \cref{fig:patches_easy_hard}, where the effect of the increasing detector noise is clearly visible.

%% file: figures/out/det-overlaps-adjusted.tex
\definecolor{mycolor1}{rgb}{0.92900,0.69400,0.12500}%
\definecolor{mycolor2}{rgb}{0.00000,0.54700,0.84100}%
\definecolor{mycolor3}{rgb}{0.95000,0.52500,0.29800}%

\definecolor{easy}{rgb}{ 0, 0.5312, 0.2148}%
\definecolor{hard}{rgb}{0.3672,0.2344,0.5977}%
\definecolor{tough}{rgb}{0.7891, 0, 0.1250}%
\begin{tikzpicture}

\begin{axis}[%
width=0.792\figW,
height=\figH,
at={(0\figW,0\figH)},
scale only axis,
xmin=0.9,
xmax=5.1,
xtick={1,2,3,4,5},
xticklabels={{2},{3},{4},{5},{6}},
xlabel={Image},
xmajorgrids,
ymin=0.4,
ymax=0.8,
ylabel={Avg. Overlap},
ymajorgrids,
axis background/.style={fill=white},
legend columns=2,
legend style={at={(1.03,0.7)},anchor=west,legend cell align=left,align=left,draw=none},
xlabel style={},ylabel style={},
]
\addplot [color=easy,dotted,line width=1.0pt,mark size=0.2pt,forget plot]
  table[row sep=crcr]{%
0.9	0.786284312270758\\
5.1	0.786284312270758\\
};
\addplot [color=hard,dotted,line width=1.0pt,mark size=0.2pt,forget plot]
  table[row sep=crcr]{%
0.9	0.627464461271266\\
5.1	0.627464461271266\\
};
\addplot [color=tough,dotted,line width=1.0pt,mark size=0.2pt,forget plot]
  table[row sep=crcr]{%
0.9	0.504982312048241\\
5.1	0.504982312048241\\
};
\addplot [color=mycolor1,dashdotted,line width=1.0pt,mark size=0.2pt,mark=x,mark options={solid},forget plot]
  table[row sep=crcr]{%
1	0.774120477779252\\
2	0.608429257100322\\
3	0.520401149801657\\
4	0.457050823688507\\
5	0.428736262003581\\
};
\addplot [color=mycolor2,dashdotted,line width=1.0pt,mark size=0.2pt,mark=x,mark options={solid},forget plot]
  table[row sep=crcr]{%
1	0.751151193968455\\
2	0.786795283136787\\
3	0.703957744021165\\
4	0.694013174578547\\
5	0.603479368961774\\
};
\addplot [color=mycolor3,dashdotted,line width=1.0pt,mark size=0.2pt,mark=x,mark options={solid},forget plot]
  table[row sep=crcr]{%
1	0.623826961770202\\
2	0.675506555171563\\
3	0.584443914441502\\
4	0.531257641648447\\
5	0.49583221150868\\
};
\addplot [color=mycolor1,solid,line width=1.0pt,mark size=0.2pt,mark=x,mark options={solid},forget plot]
  table[row sep=crcr]{%
1	0.78095774919551\\
2	0.649654058114343\\
3	0.573104671728835\\
4	0.526668939939359\\
5	0.499436461366253\\
};
\addplot [color=mycolor2,solid,line width=1.0pt,mark size=0.2pt,mark=x,mark options={solid},forget plot]
  table[row sep=crcr]{%
1	0.748306155188191\\
2	0.767093390334069\\
3	0.701425302408678\\
4	0.684463253425119\\
5	0.583302961946503\\
};
\addplot [color=mycolor3,solid,line width=1.0pt,mark size=0.2pt,mark=x,mark options={solid},forget plot]
  table[row sep=crcr]{%
1	0.647774197585477\\
2	0.692746071364048\\
3	0.593771746046195\\
4	0.540597102058379\\
5	0.496807162821794\\
};
\addplot [color=mycolor1,solid,line width=1.0pt]
  table[row sep=crcr]{%
0	0\\
};
\addlegendentry{graf};

\addplot [color=easy,dotted,line width=1.0pt]
table[row sep=crcr]{%
  0	0\\
};
\addlegendentry{\easy};

\addplot [color=mycolor2,solid,line width=1.0pt]
  table[row sep=crcr]{%
0	0\\
};
\addlegendentry{boat};

\addplot [color=hard,dotted,line width=1.0pt]
table[row sep=crcr]{%
  0	0\\
};
\addlegendentry{\hard};

\addplot [color=mycolor3,solid,line width=1.0pt]
  table[row sep=crcr]{%
0	0\\
};
\addlegendentry{wall};

\addplot [color=tough,dotted,line width=1.0pt]
table[row sep=crcr]{%
  0	0\\
};
\addlegendentry{\tough};

\addplot [color=black,dashdotted,line width=1.0pt]
  table[row sep=crcr]{%
0	0\\
};
\addlegendentry{Hes};

\addplot [color=black,solid,line width=1.0pt]
  table[row sep=crcr]{%
0	0\\
};
\addlegendentry{HesAff};

\end{axis}
\end{tikzpicture}%

%% file: hbenchmark.tex
\section{Benchmark tasks}\label{s:bench}

In this section, we define the benchmark metrics, tasks and their evaluation protocols for: patch verification, image matching and patch retrieval.

The tasks are designed to imitate typical use cases of local descriptors. Patch verification (\cref{s:patch-classification}) is based on \cite{pickingthebestdaisy} and measures the ability of a descriptor to classify whether two patches are extracted from the same measurement. Image matching (\cref{s:image-matching}), inspired by \cite{km2005pami}, tests to what extent a descriptor can correctly identify correspondences in two images. Finally, patch retrieval (\cref{s:patch-retrieval}) tests how well a descriptor can match a query patch to a pool of patches extracted from many images, including many distractors. This is a proxy to local feature based image indexing~\cite{Philbin07,paulin2015local}.

% -----------------------------------------------------------
\subsection{Evaluation metrics}\label{s:prelim}
% -----------------------------------------------------------

We first define the precision and recall evaluation metric used in \dataset. Let $\by = (y_1,\dots,y_n) \in \{-1,0,+1\}^n$ be labels of a ranked list of patches returned for a patch query, indicating negative, \emph{to ignore}, and positive match, respectively. Then \emph{precision} and \emph{recall} at rank $i$ are given by\footnote{Here $[z]_+ = \max\{0,z\}$.}
$
  P_i(\by) =
  {\sum_{k=1}^i [y_k]_+}/{\sum_{k=1}^i |y_k|}
$
and
$
  R_i(\by) =
  {\sum_{k=1}^i [y_k]_+}/{\sum_{k=1}^N [y_k]_+}
$;
the \emph{average precision} (AP) is given by 
$
  AP(\by) = \sum_{k:y_k=+1} P_k(\by) /\sum_{k=1}^N [y_k]_+. 
$
The main difference w.r.t.\ the standard definition of $PR$  is the entries that can be ignored \ie $y_i=0$ which will be used for retrieval task in section~\ref{s:patch-retrieval}. In this case, let $K \geq \sum_{k=1}^N [y_k]_+$ be the total number of positives; recall is computed as 
$
  R_i(\by;K) =
  {\sum_{k=1}^i [y_k]_+}/K
$ 
and AP as 
$
AP(\by;K) =\sum_{k:y_k=+1} P_k / K
$
which corresponds to truncated PR curves).

% -----------------------------------------------------------
\subsection{Patch verification}\label{s:patch-classification}
% -----------------------------------------------------------

In \emph{patch verification} descriptors are used to classify whether two 
patches are in correspondence or not. The benchmark starts from a list 
$\mathcal{P}=((\bx_i,\bx'_i,y_i),\ i=1,\dots,N)$ of positive and negative patch 
pairs, where $\bx_i,\bx'_i\in\mathbb{R}^{65\times 65\times 1}$ are patches and 
$y_i=\pm 1$ is their label.

The dataset is used to evaluate a matching approach $\mathcal{A}$ that, given any two patches $\bx_i,\bx_i'$, produces a confidence score $s_i\in\mathbb{R}$ that the two patches correspond. The quality of the approach is measured as the average precision of the ranked patches, namely $AP(y_{\pi_1},\dots,y_{\pi_N})$ where $\pi$ is the permutation that sorts the scores in decreasing order (i.e.\ $s_{\pi_1}\geq s_{\pi_2}\geq \dots \geq s_{\pi_n}$) to apply the formulas from section~\ref{s:prelim}.

The benchmark uses four sets of patch pairs extracted by varying the
projection noise as discussed in \cref{s:bench-data} that is \easy,
\hard or \tough as well as a set of negative pairs that are either
sampled from images within the same sequence or from different
sequences. The overall performance of the method $\mathcal{A}$ is
then computed as the mean AP for the six patch sets. In total, we
generate $2\times10^5$ positive pairs and $1\times10^6$ negative
pairs per a set.

Note that the benchmark only requires scores $s_i$ computed by the
algorithm $\mathcal{A}$; in particular, this unifies the evaluation of
a descriptor with a custom similarity metric, including a learned one.

This evaluation protocol is similar
to~\cite{pickingthebestdaisy}. However, whereas the
ROC~\cite{fawcett2004roc} is used there, AP is preferred
here~\cite{simo2015deepdesc} since the dataset is highly unbalanced,
with the vast majority ($10^6$) of patch pairs being negative. The
latter is more representative of typical matching scenarios.

% -----------------------------------------------------------
\subsection{Image matching}\label{s:image-matching}
% -----------------------------------------------------------

In \emph{image matching}, descriptors are used to match patches from a
reference image to a target one. In this task an image is a collection
of $N$ patches $L_k=(\bx_{ik},\ i=1,\dots,N)$. Consider a pair of
images $\mathcal{D}=(L_0,L_1)$, where $L_0$ is the reference and $L_1$
the target. Thus, after matching, $\bx_{i0}$ is in correspondence with
$\bx_{i1}$.

The pair $\mathcal{D}$ is used to evaluate an algorithm $\mathcal{A}$
that, given a reference patch $\bx_{i0}\in L_0$, determines the index
$\sigma_{i} \in \{1,\dots,N\}$ of the best matching patch
$\bx_{\sigma_{i}1} \in L_1$, as well as the corresponding confidence
score $s_{i}\in\mathbb{R}$. Then, the benchmark labels the assignment
$\sigma_{i}$ as $y_{i} = 2[\sigma_{i} \stackrel{\text{\tiny ?}}{=} 
i]-1$, 
and 
computes
$AP(y_{\pi_1},\dots,y_{\pi_N};N)$, where $\pi$ is the permutation that
sorts the scores in decreasing order (note that the number of positive
results is fixed to $N$; see \cref{s:prelim}).

%The benchmark extracts image pairs $\mathcal{D}$ from the image sequences in \dataset, using the first image in the sequence as reference. 
We group sequences based on whether they vary by viewpoint or
illumination and each group is instantiated with \easy, \hard and
\tough patches. The overall performance of an algorithm $\mathcal{A}$
is computed as the mean AP for all such image pairs and variants.

Note that the benchmark only requires the indexes $\sigma_{i}$ and the
scores $s_{i}$ computed by the algorithm $\mathcal{A}$ for each image
pair $\mathcal{D}$. Typically, these can be computed by extracting
patch descriptors and comparing with a similarity metric.

This evaluation protocol is designed to closely resemble the one
from~\cite{km2005pami}. A notable difference is that, since the patch
datasets are constructed in such a way that each reference patch has a
corresponding patch in each target image, the maximum recall is always
100\%.  Note also that, similarly to the verification task, the
benchmark evaluates the combined performance of the descriptor and
similarity score provided by the tested algorithm.

% -----------------------------------------------------------
\subsection{Patch retrieval}\label{s:patch-retrieval}
% -----------------------------------------------------------

In \emph{patch retrieval} descriptors are used to find patch
correspondences in a large collection of patches, a large portion of
which are distractors, extracted from confounder images. Consider a
collection $\mathcal{P} = (\bx_{0},(\bx_{i},y_{i}), i=1,\dots,N)$
consisting of a query patch $\bx_{0}$, extracted from a reference image
$L_0$, and all patches from images $L_k,k=1,\dots, K$ in the same
sequence (matching images), as well as many confounder images.

%We define a \textbf{strict} protocol, in which a patch $\bx_i$ is given a positive label $y_i=+1$ if it corresponds to the query patch $\bx_0$, and negative $y_i=-1$ otherwise. Since there is exactly one matching patch in each image $L_k$ of the same sequence, $\mathcal{D}$ contains exactly $K$ positive labels.

%We also define a \textbf{loose} protocol, in which patches $\bx_i$ that do not match the query patch $\bx_0$ but at least belong to a matching image $L_k$ are ignored ($y_i=0$). The idea is that such patches are not detrimental for the purpose of retrieving the correct image, and such innocuous errors may occur frequently in the case of repeated structures in images.

In the retrieval protocol, a patch $\bx_i$ is given a positive label
$y_i=+1$ if it corresponds to the query patch $\bx_0$, and negative
$y_i=-1$ otherwise. Since there is exactly one corresponding patch in each
image $L_k$ of the same sequence,  there are exactly $K$
positive patches in $\mathcal{D}$. However, retrieved patches $\bx_i$ that do not correspond to the query
patch $\bx_0$ but at least belong to a matching image $L_k$ are
ignored ($y_i=0$). The idea is that such patches are not detrimental
for the purpose of retrieving the correct image, and such innocuous
errors may occur frequently in the case of repeated structures in
images.

The collection $\mathcal{P}$ is used to evaluate an algorithm
$\mathcal{A}$ that assigns to each patch $\bx_i$ a confidence score
$s_i\in\mathbb{R}$ that the patch matches the query $\bx_0$. The
benchmark then returns $AP(y_{\pi_1},\dots,y_{\pi_{N}};K)$, where
$\pi$ is the permutation that sorts the scores in decreasing
order. 
%Note that, for efficiency, only the top 100 matches are considered. This is done for practical reasons, as it allows to record only the first 50 matches.

The benchmark extracts $1\times 10^4$ collections $\mathcal{P}$,
each corresponding to different query patch $\bx_0$ and its corresponding $5$ patches as well as $2\times 10^4$ distractors randomly selected from all sequences. Furthermore, there are three variants
 instantiated for \easy, \hard and \tough. 
%For the \textsc{5S} variant, we randomly draw 7 subsets, each with 5 sequences (in order to cover all 116 sequences) and sample 500 query patches from the 5 reference images. In case of \textsc{40S} variant, we select 3 random subsets of cardinality 40 and for each subset we sample 4000 query patches from the reference images. 
The overall performance of an algorithm
$\mathcal{A}$ is computed as the mean AP for all such collections and their variants.

The design of this benchmark is inspired by classical image retrieval
systems such as~\cite{Philbin07,Philbin08,paulin2015local}, which use
patches and their descriptors as entries in image indexes. A similar
evaluation may be performed by using the PhotoTourism dataset, which
includes $\sim 100\text{K}$ small sets of corresponding
patches. Unfortunately, since these small sets are not maximal, it is
not possible to know that a patch \emph{does not} have a correct correspondence without the ground truth, which makes the evaluation noisy.

%% file: experiments.tex
\section{Experimental results}\label{s:experiments}
\input{tables/det-names.tex}

\begin{table}
\caption{Basic properties of the selected descriptors. For binary descriptors, the dimensionality is in bits ($^{*}$), otherwise in number of single precision floats. The computational efficiency is measured in thousands of descriptors extracted per second.}\label{tab:descs}
\centering
{\scriptsize
  \setlength{\tabcolsep}{0.3em}
  \input{./tables/det-stats-vert-can.tex}}\textbf{}
\vspace{-1em}
\end{table}

In this section we evaluate local descriptors with the newly introduced benchmark and discuss the results in relation to the literature.

\subsection{Descriptors}

%A descriptor is a function $\mathbb{R}^{H\times H \times 3} \to \mathbb{R}^D$ mapping an image patch to a feature vector.
We evaluate the following descriptors, summarized in~\cref{tab:descs}.
We include two \textbf{baselines}: \meanstd,  $[\mu,\sigma]$ which is the average $\mu$ and standard deviation $\sigma$ of the patch, and \resize, the vector obtained by resizing the patch to $6\times 6$ pixels and normalizing it by subtracting $\mu$ and dividing by $\sigma$. 
For \textbf{SIFT-based} descriptors we include \sift~\cite{lowe1999object} and  its  variant \rootsift \cite{arandjelovic2012three}. 
%These descriptors are computed after blurring of the patch (also called differentiation scale in \cite{mikolajczyk2004scale}) as this leads to better performance. 
From the family of  \textbf{binary descriptors} we test \brief~\cite{BRIEF}, based on randomised intensity comparison, \orb~\cite{rublee2011orb}, that uses uncorrelated binary tests, and \binboost~\cite{Trzcinski2013Binboost}, where binary tests are selected using boosting. 
Finally, we evaluate several recent \textbf{deep descriptors} including the siamese variants of DeepCompare~\cite{ZagoruykoCVPR2015} (\dcsiam, \dcsiamts) with one and two stream CNN architectures for one or two patch crops, DeepDesc~\cite{simo2015deepdesc} (\deepdesc), which exploits  hard-negative mining, and the TFeat \emph{margin*} (\tfmargin) and \emph{ratio*} (\tfratio) of the TFeat descriptor~\cite{TFeat}, based on shallow convolutional networks, triplet learning constraints and fast hard negative mining. All the learning based descriptors were trained on PhotoTourism data, which is different from our new benchmark.

It has been shown in~\cite{arandjelovic2012three, bursuc2015kernel,pcaSIFT} 
that descriptor normalisation often substantially improves the performance. 
Thus, we also include post-processed variants of selected descriptors by 
applying ZCA whitening \cite[p.~299-300]{bishop1995neural} with clipped eigen 
values \cite{hua2007discriminant} followed by power law normalisation 
\cite{arandjelovic2012three} and L2 normalization. ZCA projection is computed 
on a subset of the dataset (note that ZCA is unsupervised). The threshold for 
eigen clipping is estimated for each descriptor separately to maximise its 
performance on a subset of the dataset. The normalisation is not used for 
trivial baselines and for the binary descriptors.

Table \ref{tab:descs} shows the dimensionality, size of the measurement region in pixels, and extraction time of each descriptor.  DeepCompare~\cite{ZagoruykoCVPR2015} variants have the highest dimensionality of 256 and 512, otherwise the other real value descriptors are of 128 dimensions except  \meanstd and \resize.  All binary descriptors are of 256 bits. In terms of speed, the binary descriptors \brief and \orb are 4 times faster than the most efficient CNN based  features \ie \ttb{TF-}. Other descriptors are at least an order of magnitude slower. Note that \meanstd and \resize are implemented in Matlab therefore their efficiency should be interpreted with caution.

\subsection{Results}\label{sec:results}

\begin{figure*}
  \vspace{-1em}
  \centering \footnotesize   \hspace{2.5em}
  \vspace{-0.5em}
\begin{tikzpicture}
\tikzsetnextfilename{results-legend-main}
\definecolor{easy}{rgb}{ 0, 0.5312, 0.2148}%
\definecolor{hard}{rgb}{0.3672,0.2344,0.5977}%
\definecolor{tough}{rgb}{0.7891, 0, 0.1250}%
\begin{customlegend}[legend columns=5,
legend style={align=center,draw=none},
legend cell align=left,
legend entries={\easy, \hard, \tough}]
\addlegendimage{only marks, color=easy, mark=square*}
\addlegendimage{only marks, color=hard, mark=square*}
\addlegendimage{only marks, color=tough, mark=square*}
\end{customlegend}
\end{tikzpicture}\\
  \setlength{\figH}{6cm}
  \setlength{\figW}{0.42\columnwidth}
  \begin{minipage}{0.32\linewidth} \centering \hspace{2em}
    \begin{tikzpicture}
      \tikzsetnextfilename{results-legend-verif}
      \begin{customlegend}[legend columns=2,
      legend style={align=center,draw=none},
      legend cell align=left,
      legend entries={\diffseq, \sameseq}]
      \addlegendimage{only marks, color=black, mark size=1.5pt, mark=asterisk, mark options={solid}}
      \addlegendimage{only marks, color=black, mark size=1.5pt, mark=diamond*, mark options={solid}}
      \end{customlegend}
    \end{tikzpicture}\\
    \vspace{-0.5em}
    \input{tables/verif_baselines-all.tikz}
  \end{minipage}
  \begin{minipage}{0.32\linewidth} \centering \hspace{2em}
    \begin{tikzpicture}
      \tikzsetnextfilename{results-legend-match} 
      \begin{customlegend}[legend columns=2,
      legend style={align=center,draw=none},
      legend cell align=left,
      legend entries={\viewpoint, \illum}]
      \addlegendimage{only marks, color=black, mark size=1.5pt, mark=triangle*, mark options={solid,rotate=90}}
      \addlegendimage{only marks, color=black, mark size=1.5pt, mark=x, mark options={solid}}
      \end{customlegend}
    \end{tikzpicture}\\
    \vspace{-0.5em}
    \input{tables/matching_baselines-all.tikz}
  \end{minipage}
  \begin{minipage}{0.32\linewidth} \centering  \hspace{2em}
%    \begin{tikzpicture}
%      \begin{customlegend}[legend columns=2,
%      legend style={align=center,draw=none},
%      legend cell align=left,
%      legend entries={5S, 40S}]
%      \addlegendimage{only marks, color=black, mark size=0.7pt, mark=*, mark options={solid}}
%      \addlegendimage{only marks, color=black, mark size=1.5pt, mark=*, mark options={solid}}
%      \end{customlegend}
%    \end{tikzpicture}\\
    \vspace{1.5em}
    \vspace{-0.5em}
    \input{tables/retr_patch_baselines-all.tikz}
  \end{minipage}
    \caption{Verification, matching and retrieval results.  Colour of the marker indicates \easy, \hard, and \tough noise.  The type of the marker corresponds to the variants of the experimental settings (see \cref{sec:results}). Bar is a mean of the 6 variants of each task. Dashed bar borders and $+$ indicate ZCA projected and normalised features.}\label{fig:main-results}
    \vspace{-1em}
\end{figure*}
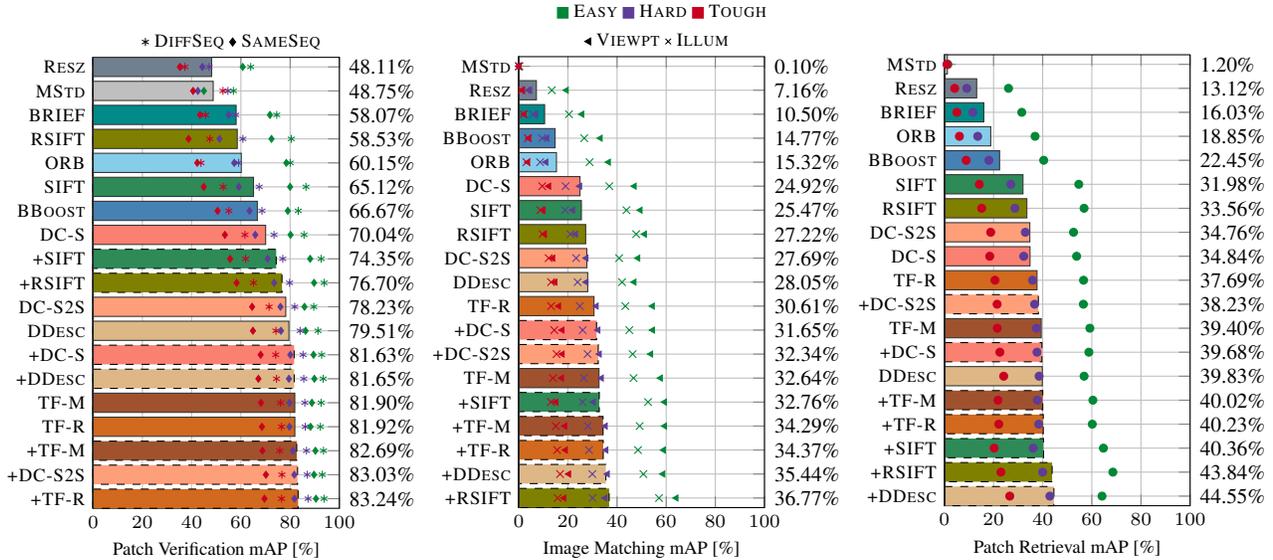

The descriptors are evaluated on three benchmark tasks: patch verification, image matching, and patch retrieval, as defined in \cref{s:bench}.
% **********For patch retrieval we report only the loose protocol (\cref{s:patch-retrieval}), but results for the strict protocol are very similar.**************
In all plots in~\cref{fig:main-results}, the colour of the marker indicates the amount of geometric noise, i.e.\ \easy, \hard, and \tough, as discussed in \cref{s:bench-data}. There are two variants of the experimental settings for each task, as explained in the discussion below, and the type of the marker corresponds to the experimental settings. The bars are the means of the six runs given by  three variants of noise with two additional settings each. Dashed bar borders and $+$ indicate ZCA projected and normalised features.

\noindent{\bf Verification.} ZCA projected and normalized +\tfratio, +\dcsiamts, 
are closely followed by other \ttb{TF-}, +\deepdesc and +\dcsiam, with slightly lower scores for post
processed \sift and binary descriptors. The post processing gives a
significant boost to \ttb{DC-} as well as \sift but a smaller improvements to \ttb{TF-} based 
descriptors. Good performance of CNN features is expected as such
descriptors are optimized together with their distance metric to perform
well in the verification task. The experiment was run for negative
pairs formed by patches from the same sequence \sameseq and from
different sequences \diffseq. The ones from \sameseq are considered
more challenging as the textures in different parts of the image are often
similar. In fact the results are consistently
lower for \sameseq. This shows that, not only the noise in positive
data poses a challenge, but the performance can also vary depending on
what source the negative examples come from.

\noindent{\bf Matching.} The ranking of descriptors changes for this
task. Although normalized +\deepdesc still performs well, surprisingly,
\prootsift comes in front of other descriptors. \ttb{+TF-}
also give good matching performance. Overall mAP scores are much lower
than for the verification task as the ratio of positive to negative
examples is significantly lower here and all the negative ones come
from the same sequence. Also the gap between \sift and deep
descriptors is narrow compared to the verification. Another
interesting observation is that the results for sequences with
photometric changes (\illum) are consistently lower than for the
viewpoint change (\viewpoint). This is different to what was observed
in evaluations on Oxford data~\cite{km2005pami}. It
seems that more progress has been made on geometric invariance in contrast to
the robustness to photometric changes. The proposed \dataset
dataset includes many sequences with extreme illumination changes.

\noindent{\bf Retrieval.}  Top performers in the retrieval scenario are the same as for matching. In particular, \sift variants are close behind +\deepdesc.  The overall performance is slightly better compared to matching which can again be explained by distractors originating from the same sequence in matching and different sequences in retrieval.

%This experiment was done for five 5S or forty sequences 40S used as distractor images. The size of the dataset seems to affect the retrieval performance more than the geometric noise as can be observed from the spread between 5S  and 40S compared to \easy and \tough.

\noindent{\bf Multitask.} There are several interesting observations across the tasks. First, the ranking of the descriptors changes, which confirms that multiple evaluation metrics are needed. Second, \sift variants, especially when followed by normalisation, perform very well. In fact, \prootsift is the second-best descriptor in both image matching and patch retrieval.  \meanstd gives good scores on verification but completely fails for matching and retrieval, as both rely on nearest neighbour matching.
Good performance on verification clearly does not generalise well to the other tasks, which much better reflect the practical applications of descriptors. This further highlights the need for using a multitask benchmark to complement training and testing on PhotoTourism, which is done in vast majority of recent papers and is similar to the verification task here. The difference in performance for \easy and \tough geometric distortions, as well as for the illumination changes, is up to 30\%, which shows there is still scope for improvement in both areas.

The performance of deep descriptors and \sift varies across the tasks although +\deepdesc~\cite{simo2015deepdesc} is close to the top scores in each category, however it is the slowest to calculate. In matching and retrieval, ZCA and normalisation bring the performance of \sift to the top level. Compared to some deep descriptors, \sift seems less robust to high degrees of geometric noise, with large spread for \easy and \tough benchmarks. This is especially evident on the patch verification task, where \sift is outperformed by most of the other descriptors for the \tough data.

The binary descriptors are outperformed by the original \sift by a large margin for the image matching and patch retrieval task in particular, which may be due to its discriminative power and better robustness to the geometric noise. The binary descriptors are competitive only for the patch verification task. However, the binary descriptors have other advantages, such as compactness and speed, so they may still be the best choice in applications where accuracy is less important than speed. Also +\ttb{TF} perform relatively well, in particular when considering their efficiency.

Post-processing normalisation, in particular square root, has a significant effect. For most of the descriptors, the normalised features perform much better than the original ones.

Finally, patch verification achieves on average much higher mAP score compared to the other tasks. This can be seen mainly from the relatively good performance of the trivial \meanstd descriptor. This confirms that patch verification task is insufficient on its own and other tasks are crucial in descriptor evaluation.

%%% Local Variables: 
%%% mode: latex
%%% TeX-master: "main"
%%% End: 

%% file: tables/det-names.tex
\newcommand{\meanstd}{{\ttb{MStd}}\xspace}
\newcommand{\resize}{{\ttb{Resz}}\xspace}
\newcommand{\sift}{{\ttb{SIFT}}\xspace}
\newcommand{\rootsift}{{\ttb{RSIFT}}\xspace}
\newcommand{\brief}{{\ttb{BRIEF}}\xspace}
\newcommand{\binboost}{{\ttb{BBoost}}\xspace}
\newcommand{\orb}{{\ttb{ORB}}\xspace}
\newcommand{\dcsiam}{{\ttb{DC-S}}\xspace}
\newcommand{\dcsiamts}{{\ttb{DC-S2S}}\xspace}
\newcommand{\deepdesc}{{\ttb{DDesc}}\xspace}
\newcommand{\tfmargin}{{\ttb{TF-M}}\xspace}
\newcommand{\tfratio}{{\ttb{TF-R}}\xspace}
\newcommand{\psift}{{\ttb{+SIFT}}\xspace}
\newcommand{\prootsift}{{\ttb{+RSIFT}}\xspace}
\newcommand{\pdcsiam}{{\ttb{+DC-S}}\xspace}
\newcommand{\pdcsiamts}{{\ttb{+DC-S2S}}\xspace}
\newcommand{\pdeepdesc}{{\ttb{+DDesc}}\xspace}
\newcommand{\ptfmargin}{{\ttb{+TF-M}}\xspace}
\newcommand{\ptfratio}{{\ttb{+TF-R}}\xspace}
\newcommand{\chance}{{\ttb{Chance}}\xspace}

%% file: tables/det-stats-vert-can.tex
\definecolor{meanstd}{rgb}{0.750,0.750,0.750}
\definecolor{resize}{rgb}{0.438,0.500,0.562}
\definecolor{sift}{rgb}{0.180,0.543,0.340}
\definecolor{rootsift}{rgb}{0.500,0.500,0.000}
\definecolor{brief}{rgb}{0.000,0.543,0.543}
\definecolor{binboost}{rgb}{0.273,0.508,0.703}
\definecolor{orb}{rgb}{0.527,0.805,0.918}
\definecolor{siam}{rgb}{0.979,0.500,0.445}
\definecolor{siam2stream}{rgb}{1.000,0.625,0.477}
\definecolor{deepdesc}{rgb}{0.867,0.719,0.527}
\definecolor{tfeat-margin-star}{rgb}{0.625,0.320,0.176}
\definecolor{tfeat-ratio-star}{rgb}{0.820,0.410,0.117}

\begin{tabular}{| l |  r  r  r  r  r  r  r  r  r  r  r  r  |}\hline
Descr. 
& \cellcolor{meanstd!75!white} \rot{\meanstd} 
& \cellcolor{resize!75!white} \rot{\resize} 
& \cellcolor{sift!75!white} \rot{\sift} 
& \cellcolor{rootsift!75!white} \rot{\rootsift} 
& \cellcolor{brief!75!white} \rot{\brief} 
& \cellcolor{binboost!75!white} \rot{\binboost} 
& \cellcolor{orb!75!white} \rot{\orb} 
& \cellcolor{siam!75!white} \rot{\dcsiam}
& \cellcolor{siam2stream!75!white} \rot{\dcsiamts} 
& \cellcolor{deepdesc!75!white} \rot{\deepdesc} 
& \cellcolor{tfeat-margin-star!75!white} \rot{\tfmargin} 
& \cellcolor{tfeat-ratio-star!75!white} \rot{\tfratio}  \\  \hline 
Dims & $2$& $36$& $128$& $128$& $^{*}256$& $^{*}256$& $^{*}256$& $256$& $512$& $128$& $128$& $128$ \\ 
Patch \;Sz & 65 & 65 & 65 & 65 & 32 & 32 & 32 & 64 & 64 & 64 & 32 & 32  \\ 
Speed \texttt{CPU} & $67$ & $3$ & $2$ & $2$ & $333$ & $2$ & $333$ & $0.3$ & $0.2$ & $0.1$ & $0.6$ & $0.6$  \\ 
Speed \texttt{GPU}  & $$ & $$ & $$ & $$ & $$ & $$ & $$ & $10$ & $5$ & $2.3$ & $83$ & $83$  \\ 
 \hline 
 \end{tabular}

%% file: tables/verif_baselines-all.tikz
\definecolor{mycolor1}{rgb}{0.82031,0.41016,0.11719}%
\definecolor{mycolor2}{rgb}{0.00000,0.53125,0.21484}%
\definecolor{mycolor3}{rgb}{0.36719,0.23438,0.59766}%
\definecolor{mycolor4}{rgb}{0.78906,0.00000,0.12500}%
\definecolor{mycolor5}{rgb}{1.00000,0.62500,0.47656}%
\definecolor{mycolor6}{rgb}{0.62500,0.32031,0.17578}%
\definecolor{mycolor7}{rgb}{0.86719,0.71875,0.52734}%
\definecolor{mycolor8}{rgb}{0.97917,0.50000,0.44531}%
\definecolor{mycolor9}{rgb}{0.17969,0.54297,0.33984}%
\definecolor{mycolor10}{rgb}{0.27344,0.50781,0.70312}%
\definecolor{mycolor11}{rgb}{0.52734,0.80469,0.91797}%
\definecolor{mycolor12}{rgb}{0.00000,0.54297,0.54297}%
\definecolor{mycolor13}{rgb}{0.43750,0.50000,0.56250}%
\begin{tikzpicture}

\begin{axis}[%
width=0.935\figW,
height=\figH,
at={(0\figW,0\figH)},
scale only axis,
clip=false,
xmin=0,
xmax=100,
xlabel={Patch Verification mAP [\%]},
y dir=reverse,
ymin=0.6,
ymax=19.4,
ytick={1,2,3,4,5,6,7,8,9,10,11,12,13,14,15,16,17,18,19},
yticklabels={{\resize},{\meanstd},{\brief},{\rootsift},{\orb},{\sift},{\binboost},{\dcsiam},{\psift},{\prootsift},{\dcsiamts},{\deepdesc},{\pdcsiam},{\pdeepdesc},{\tfmargin},{\tfratio},{\ptfmargin},{\pdcsiamts},{\ptfratio}},
axis background/.style={fill=white},
xmajorgrids,
ymajorgrids
]
\addplot[xbar, bar width=0.8, fill=mycolor1, dashed, draw=black, area legend] table[row sep=crcr] {%
83.2423874268187	19\\
};
\addplot[forget plot, color=white!15!black, dashed] table[row sep=crcr] {%
0	0.6\\
0	19.4\\
};
\node[right, align=left]
at (axis cs:101,19) {83.24\%};
\addplot [color=mycolor2, draw=none, mark size=1.5pt, mark=asterisk, mark options={solid, fill=black!60!mycolor2, mycolor2}, forget plot]
  table[row sep=crcr]{%
93.7906876377522	19\\
};
\addplot [color=mycolor3, draw=none, mark size=1.5pt, mark=asterisk, mark options={solid, fill=black!60!mycolor3, mycolor3}, forget plot]
  table[row sep=crcr]{%
87.2893482931352	19\\
};
\addplot [color=mycolor4, draw=none, mark size=1.5pt, mark=asterisk, mark options={solid, fill=black!60!mycolor4, mycolor4}, forget plot]
  table[row sep=crcr]{%
76.5375348969202	19\\
};
\addplot [color=mycolor2, draw=none, mark size=1.3pt, mark=diamond*, mark options={solid, fill=black!60!mycolor2, mycolor2}, forget plot]
  table[row sep=crcr]{%
90.4883252162073	19\\
};
\addplot [color=mycolor3, draw=none, mark size=1.3pt, mark=diamond*, mark options={solid, fill=black!60!mycolor3, mycolor3}, forget plot]
  table[row sep=crcr]{%
81.777580056856	19\\
};
\addplot [color=mycolor4, draw=none, mark size=1.3pt, mark=diamond*, mark options={solid, fill=black!60!mycolor4, mycolor4}, forget plot]
  table[row sep=crcr]{%
69.5708484600415	19\\
};
\addplot[xbar, bar width=0.8, fill=mycolor5, dashed, draw=black, area legend] table[row sep=crcr] {%
83.0278532348382	18\\
};
\addplot[forget plot, color=white!15!black, dashed] table[row sep=crcr] {%
0	0.6\\
0	19.4\\
};
\node[right, align=left]
at (axis cs:101,18) {83.03\%};
\addplot [color=mycolor2, draw=none, mark size=1.5pt, mark=asterisk, mark options={solid, fill=black!60!mycolor2, mycolor2}, forget plot]
  table[row sep=crcr]{%
93.0406482545211	18\\
};
\addplot [color=mycolor3, draw=none, mark size=1.5pt, mark=asterisk, mark options={solid, fill=black!60!mycolor3, mycolor3}, forget plot]
  table[row sep=crcr]{%
86.9247271234006	18\\
};
\addplot [color=mycolor4, draw=none, mark size=1.5pt, mark=asterisk, mark options={solid, fill=black!60!mycolor4, mycolor4}, forget plot]
  table[row sep=crcr]{%
76.6498081691663	18\\
};
\addplot [color=mycolor2, draw=none, mark size=1.3pt, mark=diamond*, mark options={solid, fill=black!60!mycolor2, mycolor2}, forget plot]
  table[row sep=crcr]{%
89.6983195123968	18\\
};
\addplot [color=mycolor3, draw=none, mark size=1.3pt, mark=diamond*, mark options={solid, fill=black!60!mycolor3, mycolor3}, forget plot]
  table[row sep=crcr]{%
81.6951330869511	18\\
};
\addplot [color=mycolor4, draw=none, mark size=1.3pt, mark=diamond*, mark options={solid, fill=black!60!mycolor4, mycolor4}, forget plot]
  table[row sep=crcr]{%
70.1584832625931	18\\
};
\addplot[xbar, bar width=0.8, fill=mycolor6, dashed, draw=black, area legend] table[row sep=crcr] {%
82.6879398171922	17\\
};
\addplot[forget plot, color=white!15!black, dashed] table[row sep=crcr] {%
0	0.6\\
0	19.4\\
};
\node[right, align=left]
at (axis cs:101,17) {82.69\%};
\addplot [color=mycolor2, draw=none, mark size=1.5pt, mark=asterisk, mark options={solid, fill=black!60!mycolor2, mycolor2}, forget plot]
  table[row sep=crcr]{%
93.5510364566089	17\\
};
\addplot [color=mycolor3, draw=none, mark size=1.5pt, mark=asterisk, mark options={solid, fill=black!60!mycolor3, mycolor3}, forget plot]
  table[row sep=crcr]{%
86.6415189853695	17\\
};
\addplot [color=mycolor4, draw=none, mark size=1.5pt, mark=asterisk, mark options={solid, fill=black!60!mycolor4, mycolor4}, forget plot]
  table[row sep=crcr]{%
75.788678659997	17\\
};
\addplot [color=mycolor2, draw=none, mark size=1.3pt, mark=diamond*, mark options={solid, fill=black!60!mycolor2, mycolor2}, forget plot]
  table[row sep=crcr]{%
90.2702646786341	17\\
};
\addplot [color=mycolor3, draw=none, mark size=1.3pt, mark=diamond*, mark options={solid, fill=black!60!mycolor3, mycolor3}, forget plot]
  table[row sep=crcr]{%
81.1245008959818	17\\
};
\addplot [color=mycolor4, draw=none, mark size=1.3pt, mark=diamond*, mark options={solid, fill=black!60!mycolor4, mycolor4}, forget plot]
  table[row sep=crcr]{%
68.7516392265622	17\\
};
\addplot[xbar, bar width=0.8, fill=mycolor1, draw=white!20!black, area legend] table[row sep=crcr] {%
81.919575763	16\\
};
\addplot[forget plot, color=white!15!black] table[row sep=crcr] {%
0	0.6\\
0	19.4\\
};
\node[right, align=left]
at (axis cs:101,16) {81.92\%};
\addplot [color=mycolor2, draw=none, mark size=1.5pt, mark=asterisk, mark options={solid, fill=black!60!mycolor2, mycolor2}, forget plot]
  table[row sep=crcr]{%
92.2905160221755	16\\
};
\addplot [color=mycolor3, draw=none, mark size=1.5pt, mark=asterisk, mark options={solid, fill=black!60!mycolor3, mycolor3}, forget plot]
  table[row sep=crcr]{%
86.1423863096678	16\\
};
\addplot [color=mycolor4, draw=none, mark size=1.5pt, mark=asterisk, mark options={solid, fill=black!60!mycolor4, mycolor4}, forget plot]
  table[row sep=crcr]{%
76.51858011235	16\\
};
\addplot [color=mycolor2, draw=none, mark size=1.3pt, mark=diamond*, mark options={solid, fill=black!60!mycolor2, mycolor2}, forget plot]
  table[row sep=crcr]{%
88.2835379688065	16\\
};
\addplot [color=mycolor3, draw=none, mark size=1.3pt, mark=diamond*, mark options={solid, fill=black!60!mycolor3, mycolor3}, forget plot]
  table[row sep=crcr]{%
79.746745813548	16\\
};
\addplot [color=mycolor4, draw=none, mark size=1.3pt, mark=diamond*, mark options={solid, fill=black!60!mycolor4, mycolor4}, forget plot]
  table[row sep=crcr]{%
68.5356883514524	16\\
};
\addplot[xbar, bar width=0.8, fill=mycolor6, draw=white!20!black, area legend] table[row sep=crcr] {%
81.897330930613	15\\
};
\addplot[forget plot, color=white!15!black] table[row sep=crcr] {%
0	0.6\\
0	19.4\\
};
\node[right, align=left]
at (axis cs:101,15) {81.90\%};
\addplot [color=mycolor2, draw=none, mark size=1.5pt, mark=asterisk, mark options={solid, fill=black!60!mycolor2, mycolor2}, forget plot]
  table[row sep=crcr]{%
92.5818288706473	15\\
};
\addplot [color=mycolor3, draw=none, mark size=1.5pt, mark=asterisk, mark options={solid, fill=black!60!mycolor3, mycolor3}, forget plot]
  table[row sep=crcr]{%
85.9791595659841	15\\
};
\addplot [color=mycolor4, draw=none, mark size=1.5pt, mark=asterisk, mark options={solid, fill=black!60!mycolor4, mycolor4}, forget plot]
  table[row sep=crcr]{%
76.0742014549435	15\\
};
\addplot [color=mycolor2, draw=none, mark size=1.3pt, mark=diamond*, mark options={solid, fill=black!60!mycolor2, mycolor2}, forget plot]
  table[row sep=crcr]{%
88.8431219017452	15\\
};
\addplot [color=mycolor3, draw=none, mark size=1.3pt, mark=diamond*, mark options={solid, fill=black!60!mycolor3, mycolor3}, forget plot]
  table[row sep=crcr]{%
79.7459691900387	15\\
};
\addplot [color=mycolor4, draw=none, mark size=1.3pt, mark=diamond*, mark options={solid, fill=black!60!mycolor4, mycolor4}, forget plot]
  table[row sep=crcr]{%
68.1597046003193	15\\
};
\addplot[xbar, bar width=0.8, fill=mycolor7, dashed, draw=black, area legend] table[row sep=crcr] {%
81.6529047404769	14\\
};
\addplot[forget plot, color=white!15!black, dashed] table[row sep=crcr] {%
0	0.6\\
0	19.4\\
};
\node[right, align=left]
at (axis cs:101,14) {81.65\%};
\addplot [color=mycolor2, draw=none, mark size=1.5pt, mark=asterisk, mark options={solid, fill=black!60!mycolor2, mycolor2}, forget plot]
  table[row sep=crcr]{%
93.4202339683072	14\\
};
\addplot [color=mycolor3, draw=none, mark size=1.5pt, mark=asterisk, mark options={solid, fill=black!60!mycolor3, mycolor3}, forget plot]
  table[row sep=crcr]{%
85.5396581515487	14\\
};
\addplot [color=mycolor4, draw=none, mark size=1.5pt, mark=asterisk, mark options={solid, fill=black!60!mycolor4, mycolor4}, forget plot]
  table[row sep=crcr]{%
74.4771034969203	14\\
};
\addplot [color=mycolor2, draw=none, mark size=1.3pt, mark=diamond*, mark options={solid, fill=black!60!mycolor2, mycolor2}, forget plot]
  table[row sep=crcr]{%
89.8452450202065	14\\
};
\addplot [color=mycolor3, draw=none, mark size=1.3pt, mark=diamond*, mark options={solid, fill=black!60!mycolor3, mycolor3}, forget plot]
  table[row sep=crcr]{%
79.5151020532503	14\\
};
\addplot [color=mycolor4, draw=none, mark size=1.3pt, mark=diamond*, mark options={solid, fill=black!60!mycolor4, mycolor4}, forget plot]
  table[row sep=crcr]{%
67.1200857526284	14\\
};
\addplot[xbar, bar width=0.8, fill=mycolor8, dashed, draw=black, area legend] table[row sep=crcr] {%
81.6317943042876	13\\
};
\addplot[forget plot, color=white!15!black, dashed] table[row sep=crcr] {%
0	0.6\\
0	19.4\\
};
\node[right, align=left]
at (axis cs:101,13) {81.63\%};
\addplot [color=mycolor2, draw=none, mark size=1.5pt, mark=asterisk, mark options={solid, fill=black!60!mycolor2, mycolor2}, forget plot]
  table[row sep=crcr]{%
92.7986769559387	13\\
};
\addplot [color=mycolor3, draw=none, mark size=1.5pt, mark=asterisk, mark options={solid, fill=black!60!mycolor3, mycolor3}, forget plot]
  table[row sep=crcr]{%
85.2359439431872	13\\
};
\addplot [color=mycolor4, draw=none, mark size=1.5pt, mark=asterisk, mark options={solid, fill=black!60!mycolor4, mycolor4}, forget plot]
  table[row sep=crcr]{%
74.1481974744175	13\\
};
\addplot [color=mycolor2, draw=none, mark size=1.3pt, mark=diamond*, mark options={solid, fill=black!60!mycolor2, mycolor2}, forget plot]
  table[row sep=crcr]{%
89.5049539556238	13\\
};
\addplot [color=mycolor3, draw=none, mark size=1.3pt, mark=diamond*, mark options={solid, fill=black!60!mycolor3, mycolor3}, forget plot]
  table[row sep=crcr]{%
80.0718398422227	13\\
};
\addplot [color=mycolor4, draw=none, mark size=1.3pt, mark=diamond*, mark options={solid, fill=black!60!mycolor4, mycolor4}, forget plot]
  table[row sep=crcr]{%
68.0311536543359	13\\
};
\addplot[xbar, bar width=0.8, fill=mycolor7, draw=white!20!black, area legend] table[row sep=crcr] {%
79.5085668508083	12\\
};
\addplot[forget plot, color=white!15!black] table[row sep=crcr] {%
0	0.6\\
0	19.4\\
};
\node[right, align=left]
at (axis cs:101,12) {79.51\%};
\addplot [color=mycolor2, draw=none, mark size=1.5pt, mark=asterisk, mark options={solid, fill=black!60!mycolor2, mycolor2}, forget plot]
  table[row sep=crcr]{%
91.3269651310997	12\\
};
\addplot [color=mycolor3, draw=none, mark size=1.5pt, mark=asterisk, mark options={solid, fill=black!60!mycolor3, mycolor3}, forget plot]
  table[row sep=crcr]{%
84.052879108711	12\\
};
\addplot [color=mycolor4, draw=none, mark size=1.5pt, mark=asterisk, mark options={solid, fill=black!60!mycolor4, mycolor4}, forget plot]
  table[row sep=crcr]{%
74.2384077022509	12\\
};
\addplot [color=mycolor2, draw=none, mark size=1.3pt, mark=diamond*, mark options={solid, fill=black!60!mycolor2, mycolor2}, forget plot]
  table[row sep=crcr]{%
86.2665940241091	12\\
};
\addplot [color=mycolor3, draw=none, mark size=1.3pt, mark=diamond*, mark options={solid, fill=black!60!mycolor3, mycolor3}, forget plot]
  table[row sep=crcr]{%
76.2771474419104	12\\
};
\addplot [color=mycolor4, draw=none, mark size=1.3pt, mark=diamond*, mark options={solid, fill=black!60!mycolor4, mycolor4}, forget plot]
  table[row sep=crcr]{%
64.8894076967686	12\\
};
\addplot[xbar, bar width=0.8, fill=mycolor5, draw=white!20!black, area legend] table[row sep=crcr] {%
78.2267833839713	11\\
};
\addplot[forget plot, color=white!15!black] table[row sep=crcr] {%
0	0.6\\
0	19.4\\
};
\node[right, align=left]
at (axis cs:101,11) {78.23\%};
\addplot [color=mycolor2, draw=none, mark size=1.5pt, mark=asterisk, mark options={solid, fill=black!60!mycolor2, mycolor2}, forget plot]
  table[row sep=crcr]{%
89.6022636019583	11\\
};
\addplot [color=mycolor3, draw=none, mark size=1.5pt, mark=asterisk, mark options={solid, fill=black!60!mycolor3, mycolor3}, forget plot]
  table[row sep=crcr]{%
81.8490371261668	11\\
};
\addplot [color=mycolor4, draw=none, mark size=1.5pt, mark=asterisk, mark options={solid, fill=black!60!mycolor4, mycolor4}, forget plot]
  table[row sep=crcr]{%
71.4579283139544	11\\
};
\addplot [color=mycolor2, draw=none, mark size=1.3pt, mark=diamond*, mark options={solid, fill=black!60!mycolor2, mycolor2}, forget plot]
  table[row sep=crcr]{%
85.8257091523709	11\\
};
\addplot [color=mycolor3, draw=none, mark size=1.3pt, mark=diamond*, mark options={solid, fill=black!60!mycolor3, mycolor3}, forget plot]
  table[row sep=crcr]{%
76.0647600266593	11\\
};
\addplot [color=mycolor4, draw=none, mark size=1.3pt, mark=diamond*, mark options={solid, fill=black!60!mycolor4, mycolor4}, forget plot]
  table[row sep=crcr]{%
64.561002082718	11\\
};
\addplot[xbar, bar width=0.8, fill=red!50!green, dashed, draw=black, area legend] table[row sep=crcr] {%
76.7030757614942	10\\
};
\addplot[forget plot, color=white!15!black, dashed] table[row sep=crcr] {%
0	0.6\\
0	19.4\\
};
\node[right, align=left]
at (axis cs:101,10) {76.70\%};
\addplot [color=mycolor2, draw=none, mark size=1.5pt, mark=asterisk, mark options={solid, fill=black!60!mycolor2, mycolor2}, forget plot]
  table[row sep=crcr]{%
93.8227628455658	10\\
};
\addplot [color=mycolor3, draw=none, mark size=1.5pt, mark=asterisk, mark options={solid, fill=black!60!mycolor3, mycolor3}, forget plot]
  table[row sep=crcr]{%
79.773142281023	10\\
};
\addplot [color=mycolor4, draw=none, mark size=1.5pt, mark=asterisk, mark options={solid, fill=black!60!mycolor4, mycolor4}, forget plot]
  table[row sep=crcr]{%
65.1241348942464	10\\
};
\addplot [color=mycolor2, draw=none, mark size=1.3pt, mark=diamond*, mark options={solid, fill=black!60!mycolor2, mycolor2}, forget plot]
  table[row sep=crcr]{%
89.7159479564517	10\\
};
\addplot [color=mycolor3, draw=none, mark size=1.3pt, mark=diamond*, mark options={solid, fill=black!60!mycolor3, mycolor3}, forget plot]
  table[row sep=crcr]{%
73.5112176733558	10\\
};
\addplot [color=mycolor4, draw=none, mark size=1.3pt, mark=diamond*, mark options={solid, fill=black!60!mycolor4, mycolor4}, forget plot]
  table[row sep=crcr]{%
58.2712489183222	10\\
};
\addplot[xbar, bar width=0.8, fill=mycolor9, dashed, draw=black, area legend] table[row sep=crcr] {%
74.3507630196642	9\\
};
\addplot[forget plot, color=white!15!black, dashed] table[row sep=crcr] {%
0	0.6\\
0	19.4\\
};
\node[right, align=left]
at (axis cs:101,9) {74.35\%};
\addplot [color=mycolor2, draw=none, mark size=1.5pt, mark=asterisk, mark options={solid, fill=black!60!mycolor2, mycolor2}, forget plot]
  table[row sep=crcr]{%
92.5846001837524	9\\
};
\addplot [color=mycolor3, draw=none, mark size=1.5pt, mark=asterisk, mark options={solid, fill=black!60!mycolor3, mycolor3}, forget plot]
  table[row sep=crcr]{%
77.0113741817108	9\\
};
\addplot [color=mycolor4, draw=none, mark size=1.5pt, mark=asterisk, mark options={solid, fill=black!60!mycolor4, mycolor4}, forget plot]
  table[row sep=crcr]{%
61.9337093102439	9\\
};
\addplot [color=mycolor2, draw=none, mark size=1.3pt, mark=diamond*, mark options={solid, fill=black!60!mycolor2, mycolor2}, forget plot]
  table[row sep=crcr]{%
88.1252851170512	9\\
};
\addplot [color=mycolor3, draw=none, mark size=1.3pt, mark=diamond*, mark options={solid, fill=black!60!mycolor3, mycolor3}, forget plot]
  table[row sep=crcr]{%
70.8418904955954	9\\
};
\addplot [color=mycolor4, draw=none, mark size=1.3pt, mark=diamond*, mark options={solid, fill=black!60!mycolor4, mycolor4}, forget plot]
  table[row sep=crcr]{%
55.6077188296316	9\\
};
\addplot[xbar, bar width=0.8, fill=mycolor8, draw=white!20!black, area legend] table[row sep=crcr] {%
70.0448538369689	8\\
};
\addplot[forget plot, color=white!15!black] table[row sep=crcr] {%
0	0.6\\
0	19.4\\
};
\node[right, align=left]
at (axis cs:101,8) {70.04\%};
\addplot [color=mycolor2, draw=none, mark size=1.5pt, mark=asterisk, mark options={solid, fill=black!60!mycolor2, mycolor2}, forget plot]
  table[row sep=crcr]{%
85.6456860014828	8\\
};
\addplot [color=mycolor3, draw=none, mark size=1.5pt, mark=asterisk, mark options={solid, fill=black!60!mycolor3, mycolor3}, forget plot]
  table[row sep=crcr]{%
73.4617594426574	8\\
};
\addplot [color=mycolor4, draw=none, mark size=1.5pt, mark=asterisk, mark options={solid, fill=black!60!mycolor4, mycolor4}, forget plot]
  table[row sep=crcr]{%
61.6382966212782	8\\
};
\addplot [color=mycolor2, draw=none, mark size=1.3pt, mark=diamond*, mark options={solid, fill=black!60!mycolor2, mycolor2}, forget plot]
  table[row sep=crcr]{%
80.1649560804981	8\\
};
\addplot [color=mycolor3, draw=none, mark size=1.3pt, mark=diamond*, mark options={solid, fill=black!60!mycolor3, mycolor3}, forget plot]
  table[row sep=crcr]{%
65.8524807570129	8\\
};
\addplot [color=mycolor4, draw=none, mark size=1.3pt, mark=diamond*, mark options={solid, fill=black!60!mycolor4, mycolor4}, forget plot]
  table[row sep=crcr]{%
53.505944118884	8\\
};
\addplot[xbar, bar width=0.8, fill=mycolor10, draw=white!20!black, area legend] table[row sep=crcr] {%
66.6730337287153	7\\
};
\addplot[forget plot, color=white!15!black] table[row sep=crcr] {%
0	0.6\\
0	19.4\\
};
\node[right, align=left]
at (axis cs:101,7) {66.67\%};
\addplot [color=mycolor2, draw=none, mark size=1.5pt, mark=asterisk, mark options={solid, fill=black!60!mycolor2, mycolor2}, forget plot]
  table[row sep=crcr]{%
83.2864346522551	7\\
};
\addplot [color=mycolor3, draw=none, mark size=1.5pt, mark=asterisk, mark options={solid, fill=black!60!mycolor3, mycolor3}, forget plot]
  table[row sep=crcr]{%
68.5584914626633	7\\
};
\addplot [color=mycolor4, draw=none, mark size=1.5pt, mark=asterisk, mark options={solid, fill=black!60!mycolor4, mycolor4}, forget plot]
  table[row sep=crcr]{%
55.0885894948942	7\\
};
\addplot [color=mycolor2, draw=none, mark size=1.3pt, mark=diamond*, mark options={solid, fill=black!60!mycolor2, mycolor2}, forget plot]
  table[row sep=crcr]{%
79.0602917494028	7\\
};
\addplot [color=mycolor3, draw=none, mark size=1.3pt, mark=diamond*, mark options={solid, fill=black!60!mycolor3, mycolor3}, forget plot]
  table[row sep=crcr]{%
63.5115739809055	7\\
};
\addplot [color=mycolor4, draw=none, mark size=1.3pt, mark=diamond*, mark options={solid, fill=black!60!mycolor4, mycolor4}, forget plot]
  table[row sep=crcr]{%
50.5328210321706	7\\
};
\addplot[xbar, bar width=0.8, fill=mycolor9, draw=white!20!black, area legend] table[row sep=crcr] {%
65.1225534360763	6\\
};
\addplot[forget plot, color=white!15!black] table[row sep=crcr] {%
0	0.6\\
0	19.4\\
};
\node[right, align=left]
at (axis cs:101,6) {65.12\%};
\addplot [color=mycolor2, draw=none, mark size=1.5pt, mark=asterisk, mark options={solid, fill=black!60!mycolor2, mycolor2}, forget plot]
  table[row sep=crcr]{%
86.4327954132451	6\\
};
\addplot [color=mycolor3, draw=none, mark size=1.5pt, mark=asterisk, mark options={solid, fill=black!60!mycolor3, mycolor3}, forget plot]
  table[row sep=crcr]{%
67.4201758972227	6\\
};
\addplot [color=mycolor4, draw=none, mark size=1.5pt, mark=asterisk, mark options={solid, fill=black!60!mycolor4, mycolor4}, forget plot]
  table[row sep=crcr]{%
52.8087386645569	6\\
};
\addplot [color=mycolor2, draw=none, mark size=1.3pt, mark=diamond*, mark options={solid, fill=black!60!mycolor2, mycolor2}, forget plot]
  table[row sep=crcr]{%
79.977534603535	6\\
};
\addplot [color=mycolor3, draw=none, mark size=1.3pt, mark=diamond*, mark options={solid, fill=black!60!mycolor3, mycolor3}, forget plot]
  table[row sep=crcr]{%
59.1781130195476	6\\
};
\addplot [color=mycolor4, draw=none, mark size=1.3pt, mark=diamond*, mark options={solid, fill=black!60!mycolor4, mycolor4}, forget plot]
  table[row sep=crcr]{%
44.9179630183507	6\\
};
\addplot[xbar, bar width=0.8, fill=mycolor11, draw=white!20!black, area legend] table[row sep=crcr] {%
60.150599429084	5\\
};
\addplot[forget plot, color=white!15!black] table[row sep=crcr] {%
0	0.6\\
0	19.4\\
};
\node[right, align=left]
at (axis cs:101,5) {60.15\%};
\addplot [color=mycolor2, draw=none, mark size=1.5pt, mark=asterisk, mark options={solid, fill=black!60!mycolor2, mycolor2}, forget plot]
  table[row sep=crcr]{%
80.0377900712492	5\\
};
\addplot [color=mycolor3, draw=none, mark size=1.5pt, mark=asterisk, mark options={solid, fill=black!60!mycolor3, mycolor3}, forget plot]
  table[row sep=crcr]{%
59.0708528794659	5\\
};
\addplot [color=mycolor4, draw=none, mark size=1.5pt, mark=asterisk, mark options={solid, fill=black!60!mycolor4, mycolor4}, forget plot]
  table[row sep=crcr]{%
43.7935821767081	5\\
};
\addplot [color=mycolor2, draw=none, mark size=1.3pt, mark=diamond*, mark options={solid, fill=black!60!mycolor2, mycolor2}, forget plot]
  table[row sep=crcr]{%
78.4371232160051	5\\
};
\addplot [color=mycolor3, draw=none, mark size=1.3pt, mark=diamond*, mark options={solid, fill=black!60!mycolor3, mycolor3}, forget plot]
  table[row sep=crcr]{%
57.2920451641329	5\\
};
\addplot [color=mycolor4, draw=none, mark size=1.3pt, mark=diamond*, mark options={solid, fill=black!60!mycolor4, mycolor4}, forget plot]
  table[row sep=crcr]{%
42.2722030669429	5\\
};
\addplot[xbar, bar width=0.8, fill=red!50!green, draw=white!20!black, area legend] table[row sep=crcr] {%
58.5259367175921	4\\
};
\addplot[forget plot, color=white!15!black] table[row sep=crcr] {%
0	0.6\\
0	19.4\\
};
\node[right, align=left]
at (axis cs:101,4) {58.53\%};
\addplot [color=mycolor2, draw=none, mark size=1.5pt, mark=asterisk, mark options={solid, fill=black!60!mycolor2, mycolor2}, forget plot]
  table[row sep=crcr]{%
80.4597130300089	4\\
};
\addplot [color=mycolor3, draw=none, mark size=1.5pt, mark=asterisk, mark options={solid, fill=black!60!mycolor3, mycolor3}, forget plot]
  table[row sep=crcr]{%
60.8053528122927	4\\
};
\addplot [color=mycolor4, draw=none, mark size=1.5pt, mark=asterisk, mark options={solid, fill=black!60!mycolor4, mycolor4}, forget plot]
  table[row sep=crcr]{%
47.3614583271403	4\\
};
\addplot [color=mycolor2, draw=none, mark size=1.3pt, mark=diamond*, mark options={solid, fill=black!60!mycolor2, mycolor2}, forget plot]
  table[row sep=crcr]{%
72.4352926858799	4\\
};
\addplot [color=mycolor3, draw=none, mark size=1.3pt, mark=diamond*, mark options={solid, fill=black!60!mycolor3, mycolor3}, forget plot]
  table[row sep=crcr]{%
51.3403560127682	4\\
};
\addplot [color=mycolor4, draw=none, mark size=1.3pt, mark=diamond*, mark options={solid, fill=black!60!mycolor4, mycolor4}, forget plot]
  table[row sep=crcr]{%
38.7534474374623	4\\
};
\addplot[xbar, bar width=0.8, fill=mycolor12, draw=white!20!black, area legend] table[row sep=crcr] {%
58.0654591051019	3\\
};
\addplot[forget plot, color=white!15!black] table[row sep=crcr] {%
0	0.6\\
0	19.4\\
};
\node[right, align=left]
at (axis cs:101,3) {58.07\%};
\addplot [color=mycolor2, draw=none, mark size=1.5pt, mark=asterisk, mark options={solid, fill=black!60!mycolor2, mycolor2}, forget plot]
  table[row sep=crcr]{%
74.5599202733735	3\\
};
\addplot [color=mycolor3, draw=none, mark size=1.5pt, mark=asterisk, mark options={solid, fill=black!60!mycolor3, mycolor3}, forget plot]
  table[row sep=crcr]{%
57.8975630464883	3\\
};
\addplot [color=mycolor4, draw=none, mark size=1.5pt, mark=asterisk, mark options={solid, fill=black!60!mycolor4, mycolor4}, forget plot]
  table[row sep=crcr]{%
45.6684688698494	3\\
};
\addplot [color=mycolor2, draw=none, mark size=1.3pt, mark=diamond*, mark options={solid, fill=black!60!mycolor2, mycolor2}, forget plot]
  table[row sep=crcr]{%
71.7965347987348	3\\
};
\addplot [color=mycolor3, draw=none, mark size=1.3pt, mark=diamond*, mark options={solid, fill=black!60!mycolor3, mycolor3}, forget plot]
  table[row sep=crcr]{%
55.0650555136083	3\\
};
\addplot [color=mycolor4, draw=none, mark size=1.3pt, mark=diamond*, mark options={solid, fill=black!60!mycolor4, mycolor4}, forget plot]
  table[row sep=crcr]{%
43.4052121285573	3\\
};
\addplot[xbar, bar width=0.8, fill=lightgray, draw=white!20!black, area legend] table[row sep=crcr] {%
48.752683706746	2\\
};
\addplot[forget plot, color=white!15!black] table[row sep=crcr] {%
0	0.6\\
0	19.4\\
};
\node[right, align=left]
at (axis cs:101,2) {48.75\%};
\addplot [color=mycolor2, draw=none, mark size=1.5pt, mark=asterisk, mark options={solid, fill=black!60!mycolor2, mycolor2}, forget plot]
  table[row sep=crcr]{%
56.9646574944832	2\\
};
\addplot [color=mycolor3, draw=none, mark size=1.5pt, mark=asterisk, mark options={solid, fill=black!60!mycolor3, mycolor3}, forget plot]
  table[row sep=crcr]{%
54.7338985076097	2\\
};
\addplot [color=mycolor4, draw=none, mark size=1.5pt, mark=asterisk, mark options={solid, fill=black!60!mycolor4, mycolor4}, forget plot]
  table[row sep=crcr]{%
52.7300393119212	2\\
};
\addplot [color=mycolor2, draw=none, mark size=1.3pt, mark=diamond*, mark options={solid, fill=black!60!mycolor2, mycolor2}, forget plot]
  table[row sep=crcr]{%
44.9736770234219	2\\
};
\addplot [color=mycolor3, draw=none, mark size=1.3pt, mark=diamond*, mark options={solid, fill=black!60!mycolor3, mycolor3}, forget plot]
  table[row sep=crcr]{%
42.5321014929306	2\\
};
\addplot [color=mycolor4, draw=none, mark size=1.3pt, mark=diamond*, mark options={solid, fill=black!60!mycolor4, mycolor4}, forget plot]
  table[row sep=crcr]{%
40.5817284101095	2\\
};
\addplot[xbar, bar width=0.8, fill=mycolor13, draw=white!20!black, area legend] table[row sep=crcr] {%
48.1137018275312	1\\
};
\addplot[forget plot, color=white!15!black] table[row sep=crcr] {%
0	0.6\\
0	19.4\\
};
\node[right, align=left]
at (axis cs:101,1) {48.11\%};
\addplot [color=mycolor2, draw=none, mark size=1.5pt, mark=asterisk, mark options={solid, fill=black!60!mycolor2, mycolor2}, forget plot]
  table[row sep=crcr]{%
63.984646282649	1\\
};
\addplot [color=mycolor3, draw=none, mark size=1.5pt, mark=asterisk, mark options={solid, fill=black!60!mycolor3, mycolor3}, forget plot]
  table[row sep=crcr]{%
47.0995787897001	1\\
};
\addplot [color=mycolor4, draw=none, mark size=1.5pt, mark=asterisk, mark options={solid, fill=black!60!mycolor4, mycolor4}, forget plot]
  table[row sep=crcr]{%
37.2726616076794	1\\
};
\addplot [color=mycolor2, draw=none, mark size=1.3pt, mark=diamond*, mark options={solid, fill=black!60!mycolor2, mycolor2}, forget plot]
  table[row sep=crcr]{%
60.7519742818393	1\\
};
\addplot [color=mycolor3, draw=none, mark size=1.3pt, mark=diamond*, mark options={solid, fill=black!60!mycolor3, mycolor3}, forget plot]
  table[row sep=crcr]{%
44.3539689799132	1\\
};
\addplot [color=mycolor4, draw=none, mark size=1.3pt, mark=diamond*, mark options={solid, fill=black!60!mycolor4, mycolor4}, forget plot]
  table[row sep=crcr]{%
35.2193810234063	1\\
};
\end{axis}
\end{tikzpicture}%

%% file: tables/matching_baselines-all.tikz
\definecolor{mycolor1}{rgb}{0.00000,0.53125,0.21484}%
\definecolor{mycolor2}{rgb}{0.36719,0.23438,0.59766}%
\definecolor{mycolor3}{rgb}{0.78906,0.00000,0.12500}%
\definecolor{mycolor4}{rgb}{0.86719,0.71875,0.52734}%
\definecolor{mycolor5}{rgb}{0.82031,0.41016,0.11719}%
\definecolor{mycolor6}{rgb}{0.62500,0.32031,0.17578}%
\definecolor{mycolor7}{rgb}{0.17969,0.54297,0.33984}%
\definecolor{mycolor8}{rgb}{1.00000,0.62500,0.47656}%
\definecolor{mycolor9}{rgb}{0.97917,0.50000,0.44531}%
\definecolor{mycolor10}{rgb}{0.52734,0.80469,0.91797}%
\definecolor{mycolor11}{rgb}{0.27344,0.50781,0.70312}%
\definecolor{mycolor12}{rgb}{0.00000,0.54297,0.54297}%
\definecolor{mycolor13}{rgb}{0.43750,0.50000,0.56250}%
\begin{tikzpicture}

\begin{axis}[%
width=0.933\figW,
height=\figH,
at={(0\figW,0\figH)},
scale only axis,
clip=false,
xmin=0,
xmax=100,
xlabel={Image Matching mAP [\%]},
y dir=reverse,
ymin=0.6,
ymax=19.4,
ytick={1,2,3,4,5,6,7,8,9,10,11,12,13,14,15,16,17,18,19},
yticklabels={{\meanstd},{\resize},{\brief},{\binboost},{\orb},{\dcsiam},{\sift},{\rootsift},{\dcsiamts},{\deepdesc},{\tfratio},{\pdcsiam},{\pdcsiamts},{\tfmargin},{\psift},{\ptfmargin},{\ptfratio},{\pdeepdesc},{\prootsift}},
axis background/.style={fill=white},
xmajorgrids,
ymajorgrids
]
\addplot[xbar, bar width=0.8, fill=red!50!green, dashed, draw=black, area legend] table[row sep=crcr] {%
36.7680340291851	19\\
};
\addplot[forget plot, color=white!15!black, dashed] table[row sep=crcr] {%
0	0.6\\
0	19.4\\
};
\node[right, align=left]
at (axis cs:101,19) {36.77\%};
\addplot [color=mycolor1, draw=none, mark size=2.0pt, mark=x, mark options={solid, fill=black!60!mycolor1, mycolor1}, forget plot]
  table[row sep=crcr]{%
57.0694055671248	19\\
};
\addplot [color=mycolor1, draw=none, mark size=1.3pt, mark=triangle*, mark options={solid, rotate=90, fill=black!60!mycolor1, mycolor1}, forget plot]
  table[row sep=crcr]{%
64.0768938227633	19\\
};
\addplot [color=mycolor2, draw=none, mark size=2.0pt, mark=x, mark options={solid, fill=black!60!mycolor2, mycolor2}, forget plot]
  table[row sep=crcr]{%
30.0239370146971	19\\
};
\addplot [color=mycolor2, draw=none, mark size=1.3pt, mark=triangle*, mark options={solid, rotate=90, fill=black!60!mycolor2, mycolor2}, forget plot]
  table[row sep=crcr]{%
35.1417635729215	19\\
};
\addplot [color=mycolor3, draw=none, mark size=2.0pt, mark=x, mark options={solid, fill=black!60!mycolor3, mycolor3}, forget plot]
  table[row sep=crcr]{%
15.8250989121685	19\\
};
\addplot [color=mycolor3, draw=none, mark size=1.3pt, mark=triangle*, mark options={solid, rotate=90, fill=black!60!mycolor3, mycolor3}, forget plot]
  table[row sep=crcr]{%
18.4711052854353	19\\
};
\addplot[xbar, bar width=0.8, fill=mycolor4, dashed, draw=black, area legend] table[row sep=crcr] {%
35.4363002390727	18\\
};
\addplot[forget plot, color=white!15!black, dashed] table[row sep=crcr] {%
0	0.6\\
0	19.4\\
};
\node[right, align=left]
at (axis cs:101,18) {35.44\%};
\addplot [color=mycolor1, draw=none, mark size=2.0pt, mark=x, mark options={solid, fill=black!60!mycolor1, mycolor1}, forget plot]
  table[row sep=crcr]{%
50.7143668199376	18\\
};
\addplot [color=mycolor1, draw=none, mark size=1.3pt, mark=triangle*, mark options={solid, rotate=90, fill=black!60!mycolor1, mycolor1}, forget plot]
  table[row sep=crcr]{%
58.6961297288927	18\\
};
\addplot [color=mycolor2, draw=none, mark size=2.0pt, mark=x, mark options={solid, fill=black!60!mycolor2, mycolor2}, forget plot]
  table[row sep=crcr]{%
29.9336166903843	18\\
};
\addplot [color=mycolor2, draw=none, mark size=1.3pt, mark=triangle*, mark options={solid, rotate=90, fill=black!60!mycolor2, mycolor2}, forget plot]
  table[row sep=crcr]{%
36.1111886713926	18\\
};
\addplot [color=mycolor3, draw=none, mark size=2.0pt, mark=x, mark options={solid, fill=black!60!mycolor3, mycolor3}, forget plot]
  table[row sep=crcr]{%
16.809489741828	18\\
};
\addplot [color=mycolor3, draw=none, mark size=1.3pt, mark=triangle*, mark options={solid, rotate=90, fill=black!60!mycolor3, mycolor3}, forget plot]
  table[row sep=crcr]{%
20.3530097820011	18\\
};
\addplot[xbar, bar width=0.8, fill=mycolor5, dashed, draw=black, area legend] table[row sep=crcr] {%
34.3719182106759	17\\
};
\addplot[forget plot, color=white!15!black, dashed] table[row sep=crcr] {%
0	0.6\\
0	19.4\\
};
\node[right, align=left]
at (axis cs:101,17) {34.37\%};
\addplot [color=mycolor1, draw=none, mark size=2.0pt, mark=x, mark options={solid, fill=black!60!mycolor1, mycolor1}, forget plot]
  table[row sep=crcr]{%
48.5162590547697	17\\
};
\addplot [color=mycolor1, draw=none, mark size=1.3pt, mark=triangle*, mark options={solid, rotate=90, fill=black!60!mycolor1, mycolor1}, forget plot]
  table[row sep=crcr]{%
58.9403683612538	17\\
};
\addplot [color=mycolor2, draw=none, mark size=2.0pt, mark=x, mark options={solid, fill=black!60!mycolor2, mycolor2}, forget plot]
  table[row sep=crcr]{%
28.5995613486284	17\\
};
\addplot [color=mycolor2, draw=none, mark size=1.3pt, mark=triangle*, mark options={solid, rotate=90, fill=black!60!mycolor2, mycolor2}, forget plot]
  table[row sep=crcr]{%
35.5245253350888	17\\
};
\addplot [color=mycolor3, draw=none, mark size=2.0pt, mark=x, mark options={solid, fill=black!60!mycolor3, mycolor3}, forget plot]
  table[row sep=crcr]{%
15.6122879562858	17\\
};
\addplot [color=mycolor3, draw=none, mark size=1.3pt, mark=triangle*, mark options={solid, rotate=90, fill=black!60!mycolor3, mycolor3}, forget plot]
  table[row sep=crcr]{%
19.038507208029	17\\
};
\addplot[xbar, bar width=0.8, fill=mycolor6, dashed, draw=black, area legend] table[row sep=crcr] {%
34.294664091878	16\\
};
\addplot[forget plot, color=white!15!black, dashed] table[row sep=crcr] {%
0	0.6\\
0	19.4\\
};
\node[right, align=left]
at (axis cs:101,16) {34.29\%};
\addplot [color=mycolor1, draw=none, mark size=2.0pt, mark=x, mark options={solid, fill=black!60!mycolor1, mycolor1}, forget plot]
  table[row sep=crcr]{%
49.1896909729659	16\\
};
\addplot [color=mycolor1, draw=none, mark size=1.3pt, mark=triangle*, mark options={solid, rotate=90, fill=black!60!mycolor1, mycolor1}, forget plot]
  table[row sep=crcr]{%
59.393972879754	16\\
};
\addplot [color=mycolor2, draw=none, mark size=2.0pt, mark=x, mark options={solid, fill=black!60!mycolor2, mycolor2}, forget plot]
  table[row sep=crcr]{%
28.0886931169706	16\\
};
\addplot [color=mycolor2, draw=none, mark size=1.3pt, mark=triangle*, mark options={solid, rotate=90, fill=black!60!mycolor2, mycolor2}, forget plot]
  table[row sep=crcr]{%
35.2705053592811	16\\
};
\addplot [color=mycolor3, draw=none, mark size=2.0pt, mark=x, mark options={solid, fill=black!60!mycolor3, mycolor3}, forget plot]
  table[row sep=crcr]{%
15.0657788883787	16\\
};
\addplot [color=mycolor3, draw=none, mark size=1.3pt, mark=triangle*, mark options={solid, rotate=90, fill=black!60!mycolor3, mycolor3}, forget plot]
  table[row sep=crcr]{%
18.7593433339175	16\\
};
\addplot[xbar, bar width=0.8, fill=mycolor7, dashed, draw=black, area legend] table[row sep=crcr] {%
32.7605270745556	15\\
};
\addplot[forget plot, color=white!15!black, dashed] table[row sep=crcr] {%
0	0.6\\
0	19.4\\
};
\node[right, align=left]
at (axis cs:101,15) {32.76\%};
\addplot [color=mycolor1, draw=none, mark size=2.0pt, mark=x, mark options={solid, fill=black!60!mycolor1, mycolor1}, forget plot]
  table[row sep=crcr]{%
52.6041948857985	15\\
};
\addplot [color=mycolor1, draw=none, mark size=1.3pt, mark=triangle*, mark options={solid, rotate=90, fill=black!60!mycolor1, mycolor1}, forget plot]
  table[row sep=crcr]{%
59.3655301872247	15\\
};
\addplot [color=mycolor2, draw=none, mark size=2.0pt, mark=x, mark options={solid, fill=black!60!mycolor2, mycolor2}, forget plot]
  table[row sep=crcr]{%
25.9167198198651	15\\
};
\addplot [color=mycolor2, draw=none, mark size=1.3pt, mark=triangle*, mark options={solid, rotate=90, fill=black!60!mycolor2, mycolor2}, forget plot]
  table[row sep=crcr]{%
30.4293972663001	15\\
};
\addplot [color=mycolor3, draw=none, mark size=2.0pt, mark=x, mark options={solid, fill=black!60!mycolor3, mycolor3}, forget plot]
  table[row sep=crcr]{%
13.1003870902628	15\\
};
\addplot [color=mycolor3, draw=none, mark size=1.3pt, mark=triangle*, mark options={solid, rotate=90, fill=black!60!mycolor3, mycolor3}, forget plot]
  table[row sep=crcr]{%
15.1469331978821	15\\
};
\addplot[xbar, bar width=0.8, fill=mycolor6, draw=white!20!black, area legend] table[row sep=crcr] {%
32.6447117568601	14\\
};
\addplot[forget plot, color=white!15!black] table[row sep=crcr] {%
0	0.6\\
0	19.4\\
};
\node[right, align=left]
at (axis cs:101,14) {32.64\%};
\addplot [color=mycolor1, draw=none, mark size=2.0pt, mark=x, mark options={solid, fill=black!60!mycolor1, mycolor1}, forget plot]
  table[row sep=crcr]{%
46.7024760142034	14\\
};
\addplot [color=mycolor1, draw=none, mark size=1.3pt, mark=triangle*, mark options={solid, rotate=90, fill=black!60!mycolor1, mycolor1}, forget plot]
  table[row sep=crcr]{%
57.6819933818794	14\\
};
\addplot [color=mycolor2, draw=none, mark size=2.0pt, mark=x, mark options={solid, fill=black!60!mycolor2, mycolor2}, forget plot]
  table[row sep=crcr]{%
26.432081136281	14\\
};
\addplot [color=mycolor2, draw=none, mark size=1.3pt, mark=triangle*, mark options={solid, rotate=90, fill=black!60!mycolor2, mycolor2}, forget plot]
  table[row sep=crcr]{%
33.6780694598851	14\\
};
\addplot [color=mycolor3, draw=none, mark size=2.0pt, mark=x, mark options={solid, fill=black!60!mycolor3, mycolor3}, forget plot]
  table[row sep=crcr]{%
13.8975408595926	14\\
};
\addplot [color=mycolor3, draw=none, mark size=1.3pt, mark=triangle*, mark options={solid, rotate=90, fill=black!60!mycolor3, mycolor3}, forget plot]
  table[row sep=crcr]{%
17.4761096893193	14\\
};
\addplot[xbar, bar width=0.8, fill=mycolor8, dashed, draw=black, area legend] table[row sep=crcr] {%
32.3358566276707	13\\
};
\addplot[forget plot, color=white!15!black, dashed] table[row sep=crcr] {%
0	0.6\\
0	19.4\\
};
\node[right, align=left]
at (axis cs:101,13) {32.34\%};
\addplot [color=mycolor1, draw=none, mark size=2.0pt, mark=x, mark options={solid, fill=black!60!mycolor1, mycolor1}, forget plot]
  table[row sep=crcr]{%
46.3404763522916	13\\
};
\addplot [color=mycolor1, draw=none, mark size=1.3pt, mark=triangle*, mark options={solid, rotate=90, fill=black!60!mycolor1, mycolor1}, forget plot]
  table[row sep=crcr]{%
53.752865932259	13\\
};
\addplot [color=mycolor2, draw=none, mark size=2.0pt, mark=x, mark options={solid, fill=black!60!mycolor2, mycolor2}, forget plot]
  table[row sep=crcr]{%
27.985474714736	13\\
};
\addplot [color=mycolor2, draw=none, mark size=1.3pt, mark=triangle*, mark options={solid, rotate=90, fill=black!60!mycolor2, mycolor2}, forget plot]
  table[row sep=crcr]{%
32.8324560907618	13\\
};
\addplot [color=mycolor3, draw=none, mark size=2.0pt, mark=x, mark options={solid, fill=black!60!mycolor3, mycolor3}, forget plot]
  table[row sep=crcr]{%
15.3633708774757	13\\
};
\addplot [color=mycolor3, draw=none, mark size=1.3pt, mark=triangle*, mark options={solid, rotate=90, fill=black!60!mycolor3, mycolor3}, forget plot]
  table[row sep=crcr]{%
17.7404957985002	13\\
};
\addplot[xbar, bar width=0.8, fill=mycolor9, dashed, draw=black, area legend] table[row sep=crcr] {%
31.6455491687258	12\\
};
\addplot[forget plot, color=white!15!black, dashed] table[row sep=crcr] {%
0	0.6\\
0	19.4\\
};
\node[right, align=left]
at (axis cs:101,12) {31.65\%};
\addplot [color=mycolor1, draw=none, mark size=2.0pt, mark=x, mark options={solid, fill=black!60!mycolor1, mycolor1}, forget plot]
  table[row sep=crcr]{%
44.9892041883395	12\\
};
\addplot [color=mycolor1, draw=none, mark size=1.3pt, mark=triangle*, mark options={solid, rotate=90, fill=black!60!mycolor1, mycolor1}, forget plot]
  table[row sep=crcr]{%
54.4832354300386	12\\
};
\addplot [color=mycolor2, draw=none, mark size=2.0pt, mark=x, mark options={solid, fill=black!60!mycolor2, mycolor2}, forget plot]
  table[row sep=crcr]{%
25.9914575855059	12\\
};
\addplot [color=mycolor2, draw=none, mark size=1.3pt, mark=triangle*, mark options={solid, rotate=90, fill=black!60!mycolor2, mycolor2}, forget plot]
  table[row sep=crcr]{%
32.2940195381653	12\\
};
\addplot [color=mycolor3, draw=none, mark size=2.0pt, mark=x, mark options={solid, fill=black!60!mycolor3, mycolor3}, forget plot]
  table[row sep=crcr]{%
14.4382158432782	12\\
};
\addplot [color=mycolor3, draw=none, mark size=1.3pt, mark=triangle*, mark options={solid, rotate=90, fill=black!60!mycolor3, mycolor3}, forget plot]
  table[row sep=crcr]{%
17.6771624270271	12\\
};
\addplot[xbar, bar width=0.8, fill=mycolor5, draw=white!20!black, area legend] table[row sep=crcr] {%
30.6062124934339	11\\
};
\addplot[forget plot, color=white!15!black] table[row sep=crcr] {%
0	0.6\\
0	19.4\\
};
\node[right, align=left]
at (axis cs:101,11) {30.61\%};
\addplot [color=mycolor1, draw=none, mark size=2.0pt, mark=x, mark options={solid, fill=black!60!mycolor1, mycolor1}, forget plot]
  table[row sep=crcr]{%
43.358121323593	11\\
};
\addplot [color=mycolor1, draw=none, mark size=1.3pt, mark=triangle*, mark options={solid, rotate=90, fill=black!60!mycolor1, mycolor1}, forget plot]
  table[row sep=crcr]{%
54.422639664462	11\\
};
\addplot [color=mycolor2, draw=none, mark size=2.0pt, mark=x, mark options={solid, fill=black!60!mycolor2, mycolor2}, forget plot]
  table[row sep=crcr]{%
24.8419646860756	11\\
};
\addplot [color=mycolor2, draw=none, mark size=1.3pt, mark=triangle*, mark options={solid, rotate=90, fill=black!60!mycolor2, mycolor2}, forget plot]
  table[row sep=crcr]{%
31.5800632672336	11\\
};
\addplot [color=mycolor3, draw=none, mark size=2.0pt, mark=x, mark options={solid, fill=black!60!mycolor3, mycolor3}, forget plot]
  table[row sep=crcr]{%
13.1526480204117	11\\
};
\addplot [color=mycolor3, draw=none, mark size=1.3pt, mark=triangle*, mark options={solid, rotate=90, fill=black!60!mycolor3, mycolor3}, forget plot]
  table[row sep=crcr]{%
16.2818379988277	11\\
};
\addplot[xbar, bar width=0.8, fill=mycolor4, draw=white!20!black, area legend] table[row sep=crcr] {%
28.0516522248867	10\\
};
\addplot[forget plot, color=white!15!black] table[row sep=crcr] {%
0	0.6\\
0	19.4\\
};
\node[right, align=left]
at (axis cs:101,10) {28.05\%};
\addplot [color=mycolor1, draw=none, mark size=2.0pt, mark=x, mark options={solid, fill=black!60!mycolor1, mycolor1}, forget plot]
  table[row sep=crcr]{%
42.0376429514728	10\\
};
\addplot [color=mycolor1, draw=none, mark size=1.3pt, mark=triangle*, mark options={solid, rotate=90, fill=black!60!mycolor1, mycolor1}, forget plot]
  table[row sep=crcr]{%
46.9734649037756	10\\
};
\addplot [color=mycolor2, draw=none, mark size=2.0pt, mark=x, mark options={solid, fill=black!60!mycolor2, mycolor2}, forget plot]
  table[row sep=crcr]{%
23.9285321872676	10\\
};
\addplot [color=mycolor2, draw=none, mark size=1.3pt, mark=triangle*, mark options={solid, rotate=90, fill=black!60!mycolor2, mycolor2}, forget plot]
  table[row sep=crcr]{%
27.4747487020409	10\\
};
\addplot [color=mycolor3, draw=none, mark size=2.0pt, mark=x, mark options={solid, fill=black!60!mycolor3, mycolor3}, forget plot]
  table[row sep=crcr]{%
13.1370880355163	10\\
};
\addplot [color=mycolor3, draw=none, mark size=1.3pt, mark=triangle*, mark options={solid, rotate=90, fill=black!60!mycolor3, mycolor3}, forget plot]
  table[row sep=crcr]{%
14.7584365692472	10\\
};
\addplot[xbar, bar width=0.8, fill=mycolor8, draw=white!20!black, area legend] table[row sep=crcr] {%
27.6924610243467	9\\
};
\addplot[forget plot, color=white!15!black] table[row sep=crcr] {%
0	0.6\\
0	19.4\\
};
\node[right, align=left]
at (axis cs:101,9) {27.69\%};
\addplot [color=mycolor1, draw=none, mark size=2.0pt, mark=x, mark options={solid, fill=black!60!mycolor1, mycolor1}, forget plot]
  table[row sep=crcr]{%
40.8444970146086	9\\
};
\addplot [color=mycolor1, draw=none, mark size=1.3pt, mark=triangle*, mark options={solid, rotate=90, fill=black!60!mycolor1, mycolor1}, forget plot]
  table[row sep=crcr]{%
48.5092344560324	9\\
};
\addplot [color=mycolor2, draw=none, mark size=2.0pt, mark=x, mark options={solid, fill=black!60!mycolor2, mycolor2}, forget plot]
  table[row sep=crcr]{%
23.3255997072442	9\\
};
\addplot [color=mycolor2, draw=none, mark size=1.3pt, mark=triangle*, mark options={solid, rotate=90, fill=black!60!mycolor2, mycolor2}, forget plot]
  table[row sep=crcr]{%
27.4443758564284	9\\
};
\addplot [color=mycolor3, draw=none, mark size=2.0pt, mark=x, mark options={solid, fill=black!60!mycolor3, mycolor3}, forget plot]
  table[row sep=crcr]{%
12.1490059108016	9\\
};
\addplot [color=mycolor3, draw=none, mark size=1.3pt, mark=triangle*, mark options={solid, rotate=90, fill=black!60!mycolor3, mycolor3}, forget plot]
  table[row sep=crcr]{%
13.882053200965	9\\
};
\addplot[xbar, bar width=0.8, fill=red!50!green, draw=white!20!black, area legend] table[row sep=crcr] {%
27.2249298737059	8\\
};
\addplot[forget plot, color=white!15!black] table[row sep=crcr] {%
0	0.6\\
0	19.4\\
};
\node[right, align=left]
at (axis cs:101,8) {27.22\%};
\addplot [color=mycolor1, draw=none, mark size=2.0pt, mark=x, mark options={solid, fill=black!60!mycolor1, mycolor1}, forget plot]
  table[row sep=crcr]{%
47.7661604714202	8\\
};
\addplot [color=mycolor1, draw=none, mark size=1.3pt, mark=triangle*, mark options={solid, rotate=90, fill=black!60!mycolor1, mycolor1}, forget plot]
  table[row sep=crcr]{%
51.2080162855921	8\\
};
\addplot [color=mycolor2, draw=none, mark size=2.0pt, mark=x, mark options={solid, fill=black!60!mycolor2, mycolor2}, forget plot]
  table[row sep=crcr]{%
21.0011108028897	8\\
};
\addplot [color=mycolor2, draw=none, mark size=1.3pt, mark=triangle*, mark options={solid, rotate=90, fill=black!60!mycolor2, mycolor2}, forget plot]
  table[row sep=crcr]{%
23.1606686091906	8\\
};
\addplot [color=mycolor3, draw=none, mark size=2.0pt, mark=x, mark options={solid, fill=black!60!mycolor3, mycolor3}, forget plot]
  table[row sep=crcr]{%
9.68222043040179	8\\
};
\addplot [color=mycolor3, draw=none, mark size=1.3pt, mark=triangle*, mark options={solid, rotate=90, fill=black!60!mycolor3, mycolor3}, forget plot]
  table[row sep=crcr]{%
10.531402642741	8\\
};
\addplot[xbar, bar width=0.8, fill=mycolor7, draw=white!20!black, area legend] table[row sep=crcr] {%
25.4711896414266	7\\
};
\addplot[forget plot, color=white!15!black] table[row sep=crcr] {%
0	0.6\\
0	19.4\\
};
\node[right, align=left]
at (axis cs:101,7) {25.47\%};
\addplot [color=mycolor1, draw=none, mark size=2.0pt, mark=x, mark options={solid, fill=black!60!mycolor1, mycolor1}, forget plot]
  table[row sep=crcr]{%
43.8118067975538	7\\
};
\addplot [color=mycolor1, draw=none, mark size=1.3pt, mark=triangle*, mark options={solid, rotate=90, fill=black!60!mycolor1, mycolor1}, forget plot]
  table[row sep=crcr]{%
49.4502348037078	7\\
};
\addplot [color=mycolor2, draw=none, mark size=2.0pt, mark=x, mark options={solid, fill=black!60!mycolor2, mycolor2}, forget plot]
  table[row sep=crcr]{%
18.8334503781831	7\\
};
\addplot [color=mycolor2, draw=none, mark size=1.3pt, mark=triangle*, mark options={solid, rotate=90, fill=black!60!mycolor2, mycolor2}, forget plot]
  table[row sep=crcr]{%
21.9675577721178	7\\
};
\addplot [color=mycolor3, draw=none, mark size=2.0pt, mark=x, mark options={solid, fill=black!60!mycolor3, mycolor3}, forget plot]
  table[row sep=crcr]{%
8.77810032835944	7\\
};
\addplot [color=mycolor3, draw=none, mark size=1.3pt, mark=triangle*, mark options={solid, rotate=90, fill=black!60!mycolor3, mycolor3}, forget plot]
  table[row sep=crcr]{%
9.98598776863759	7\\
};
\addplot[xbar, bar width=0.8, fill=mycolor9, draw=white!20!black, area legend] table[row sep=crcr] {%
24.9223221016742	6\\
};
\addplot[forget plot, color=white!15!black] table[row sep=crcr] {%
0	0.6\\
0	19.4\\
};
\node[right, align=left]
at (axis cs:101,6) {24.92\%};
\addplot [color=mycolor1, draw=none, mark size=2.0pt, mark=x, mark options={solid, fill=black!60!mycolor1, mycolor1}, forget plot]
  table[row sep=crcr]{%
36.8432612069869	6\\
};
\addplot [color=mycolor1, draw=none, mark size=1.3pt, mark=triangle*, mark options={solid, rotate=90, fill=black!60!mycolor1, mycolor1}, forget plot]
  table[row sep=crcr]{%
47.0486195633751	6\\
};
\addplot [color=mycolor2, draw=none, mark size=2.0pt, mark=x, mark options={solid, fill=black!60!mycolor2, mycolor2}, forget plot]
  table[row sep=crcr]{%
19.033484971501	6\\
};
\addplot [color=mycolor2, draw=none, mark size=1.3pt, mark=triangle*, mark options={solid, rotate=90, fill=black!60!mycolor2, mycolor2}, forget plot]
  table[row sep=crcr]{%
24.8414917486505	6\\
};
\addplot [color=mycolor3, draw=none, mark size=2.0pt, mark=x, mark options={solid, fill=black!60!mycolor3, mycolor3}, forget plot]
  table[row sep=crcr]{%
9.52206688441754	6\\
};
\addplot [color=mycolor3, draw=none, mark size=1.3pt, mark=triangle*, mark options={solid, rotate=90, fill=black!60!mycolor3, mycolor3}, forget plot]
  table[row sep=crcr]{%
12.2450082351139	6\\
};
\addplot[xbar, bar width=0.8, fill=mycolor10, draw=white!20!black, area legend] table[row sep=crcr] {%
15.3249086480103	5\\
};
\addplot[forget plot, color=white!15!black] table[row sep=crcr] {%
0	0.6\\
0	19.4\\
};
\node[right, align=left]
at (axis cs:101,5) {15.32\%};
\addplot [color=mycolor1, draw=none, mark size=2.0pt, mark=x, mark options={solid, fill=black!60!mycolor1, mycolor1}, forget plot]
  table[row sep=crcr]{%
28.8096571238051	5\\
};
\addplot [color=mycolor1, draw=none, mark size=1.3pt, mark=triangle*, mark options={solid, rotate=90, fill=black!60!mycolor1, mycolor1}, forget plot]
  table[row sep=crcr]{%
36.5416560917409	5\\
};
\addplot [color=mycolor2, draw=none, mark size=2.0pt, mark=x, mark options={solid, fill=black!60!mycolor2, mycolor2}, forget plot]
  table[row sep=crcr]{%
8.67619268703368	5\\
};
\addplot [color=mycolor2, draw=none, mark size=1.3pt, mark=triangle*, mark options={solid, rotate=90, fill=black!60!mycolor2, mycolor2}, forget plot]
  table[row sep=crcr]{%
11.098883796147	5\\
};
\addplot [color=mycolor3, draw=none, mark size=2.0pt, mark=x, mark options={solid, fill=black!60!mycolor3, mycolor3}, forget plot]
  table[row sep=crcr]{%
3.12112920316641	5\\
};
\addplot [color=mycolor3, draw=none, mark size=1.3pt, mark=triangle*, mark options={solid, rotate=90, fill=black!60!mycolor3, mycolor3}, forget plot]
  table[row sep=crcr]{%
3.70193298616887	5\\
};
\addplot[xbar, bar width=0.8, fill=mycolor11, draw=white!20!black, area legend] table[row sep=crcr] {%
14.773220264826	4\\
};
\addplot[forget plot, color=white!15!black] table[row sep=crcr] {%
0	0.6\\
0	19.4\\
};
\node[right, align=left]
at (axis cs:101,4) {14.77\%};
\addplot [color=mycolor1, draw=none, mark size=2.0pt, mark=x, mark options={solid, fill=black!60!mycolor1, mycolor1}, forget plot]
  table[row sep=crcr]{%
26.6365092763143	4\\
};
\addplot [color=mycolor1, draw=none, mark size=1.3pt, mark=triangle*, mark options={solid, rotate=90, fill=black!60!mycolor1, mycolor1}, forget plot]
  table[row sep=crcr]{%
33.2197488228928	4\\
};
\addplot [color=mycolor2, draw=none, mark size=2.0pt, mark=x, mark options={solid, fill=black!60!mycolor2, mycolor2}, forget plot]
  table[row sep=crcr]{%
9.50072660756408	4\\
};
\addplot [color=mycolor2, draw=none, mark size=1.3pt, mark=triangle*, mark options={solid, rotate=90, fill=black!60!mycolor2, mycolor2}, forget plot]
  table[row sep=crcr]{%
11.4213295392022	4\\
};
\addplot [color=mycolor3, draw=none, mark size=2.0pt, mark=x, mark options={solid, fill=black!60!mycolor3, mycolor3}, forget plot]
  table[row sep=crcr]{%
3.74036446974219	4\\
};
\addplot [color=mycolor3, draw=none, mark size=1.3pt, mark=triangle*, mark options={solid, rotate=90, fill=black!60!mycolor3, mycolor3}, forget plot]
  table[row sep=crcr]{%
4.1206428732403	4\\
};
\addplot[xbar, bar width=0.8, fill=mycolor12, draw=white!20!black, area legend] table[row sep=crcr] {%
10.5033858194734	3\\
};
\addplot[forget plot, color=white!15!black] table[row sep=crcr] {%
0	0.6\\
0	19.4\\
};
\node[right, align=left]
at (axis cs:101,3) {10.50\%};
\addplot [color=mycolor1, draw=none, mark size=2.0pt, mark=x, mark options={solid, fill=black!60!mycolor1, mycolor1}, forget plot]
  table[row sep=crcr]{%
20.4316640095065	3\\
};
\addplot [color=mycolor1, draw=none, mark size=1.3pt, mark=triangle*, mark options={solid, rotate=90, fill=black!60!mycolor1, mycolor1}, forget plot]
  table[row sep=crcr]{%
25.7112093407131	3\\
};
\addplot [color=mycolor2, draw=none, mark size=2.0pt, mark=x, mark options={solid, fill=black!60!mycolor2, mycolor2}, forget plot]
  table[row sep=crcr]{%
5.75081375204824	3\\
};
\addplot [color=mycolor2, draw=none, mark size=1.3pt, mark=triangle*, mark options={solid, rotate=90, fill=black!60!mycolor2, mycolor2}, forget plot]
  table[row sep=crcr]{%
6.90372760186882	3\\
};
\addplot [color=mycolor3, draw=none, mark size=2.0pt, mark=x, mark options={solid, fill=black!60!mycolor3, mycolor3}, forget plot]
  table[row sep=crcr]{%
1.99073117015236	3\\
};
\addplot [color=mycolor3, draw=none, mark size=1.3pt, mark=triangle*, mark options={solid, rotate=90, fill=black!60!mycolor3, mycolor3}, forget plot]
  table[row sep=crcr]{%
2.23216904255141	3\\
};
\addplot[xbar, bar width=0.8, fill=mycolor13, draw=white!20!black, area legend] table[row sep=crcr] {%
7.16195493961423	2\\
};
\addplot[forget plot, color=white!15!black] table[row sep=crcr] {%
0	0.6\\
0	19.4\\
};
\node[right, align=left]
at (axis cs:101,2) {7.16\%};
\addplot [color=mycolor1, draw=none, mark size=2.0pt, mark=x, mark options={solid, fill=black!60!mycolor1, mycolor1}, forget plot]
  table[row sep=crcr]{%
13.4231643808122	2\\
};
\addplot [color=mycolor1, draw=none, mark size=1.3pt, mark=triangle*, mark options={solid, rotate=90, fill=black!60!mycolor1, mycolor1}, forget plot]
  table[row sep=crcr]{%
19.4125996342962	2\\
};
\addplot [color=mycolor2, draw=none, mark size=2.0pt, mark=x, mark options={solid, fill=black!60!mycolor2, mycolor2}, forget plot]
  table[row sep=crcr]{%
3.10733151106867	2\\
};
\addplot [color=mycolor2, draw=none, mark size=1.3pt, mark=triangle*, mark options={solid, rotate=90, fill=black!60!mycolor2, mycolor2}, forget plot]
  table[row sep=crcr]{%
4.53162618665791	2\\
};
\addplot [color=mycolor3, draw=none, mark size=2.0pt, mark=x, mark options={solid, fill=black!60!mycolor3, mycolor3}, forget plot]
  table[row sep=crcr]{%
1.06565527696055	2\\
};
\addplot [color=mycolor3, draw=none, mark size=1.3pt, mark=triangle*, mark options={solid, rotate=90, fill=black!60!mycolor3, mycolor3}, forget plot]
  table[row sep=crcr]{%
1.43135264788981	2\\
};
\addplot[xbar, bar width=0.8, fill=lightgray, draw=white!20!black, area legend] table[row sep=crcr] {%
0.0982106041906467	1\\
};
\addplot[forget plot, color=white!15!black] table[row sep=crcr] {%
0	0.6\\
0	19.4\\
};
\node[right, align=left]
at (axis cs:101,1) {0.10\%};
\addplot [color=mycolor1, draw=none, mark size=2.0pt, mark=x, mark options={solid, fill=black!60!mycolor1, mycolor1}, forget plot]
  table[row sep=crcr]{%
0.17945662116986	1\\
};
\addplot [color=mycolor1, draw=none, mark size=1.3pt, mark=triangle*, mark options={solid, rotate=90, fill=black!60!mycolor1, mycolor1}, forget plot]
  table[row sep=crcr]{%
0.130791244920481	1\\
};
\addplot [color=mycolor2, draw=none, mark size=2.0pt, mark=x, mark options={solid, fill=black!60!mycolor2, mycolor2}, forget plot]
  table[row sep=crcr]{%
0.0957432407476135	1\\
};
\addplot [color=mycolor2, draw=none, mark size=1.3pt, mark=triangle*, mark options={solid, rotate=90, fill=black!60!mycolor2, mycolor2}, forget plot]
  table[row sep=crcr]{%
0.0740708413793903	1\\
};
\addplot [color=mycolor3, draw=none, mark size=2.0pt, mark=x, mark options={solid, fill=black!60!mycolor3, mycolor3}, forget plot]
  table[row sep=crcr]{%
0.060008686678115	1\\
};
\addplot [color=mycolor3, draw=none, mark size=1.3pt, mark=triangle*, mark options={solid, rotate=90, fill=black!60!mycolor3, mycolor3}, forget plot]
  table[row sep=crcr]{%
0.0491929902484197	1\\
};
\end{axis}
\end{tikzpicture}%

%% file: tables/retr_patch_baselines-all.tikz
\definecolor{mycolor1}{rgb}{0.86719,0.71875,0.52734}%
\definecolor{mycolor2}{rgb}{0.00000,0.53125,0.21484}%
\definecolor{mycolor3}{rgb}{0.36719,0.23438,0.59766}%
\definecolor{mycolor4}{rgb}{0.78906,0.00000,0.12500}%
\definecolor{mycolor5}{rgb}{0.17969,0.54297,0.33984}%
\definecolor{mycolor6}{rgb}{0.82031,0.41016,0.11719}%
\definecolor{mycolor7}{rgb}{0.62500,0.32031,0.17578}%
\definecolor{mycolor8}{rgb}{0.97917,0.50000,0.44531}%
\definecolor{mycolor9}{rgb}{1.00000,0.62500,0.47656}%
\definecolor{mycolor10}{rgb}{0.27344,0.50781,0.70312}%
\definecolor{mycolor11}{rgb}{0.52734,0.80469,0.91797}%
\definecolor{mycolor12}{rgb}{0.00000,0.54297,0.54297}%
\definecolor{mycolor13}{rgb}{0.43750,0.50000,0.56250}%
\begin{tikzpicture}

\begin{axis}[%
width=0.933\figW,
height=\figH,
at={(0\figW,0\figH)},
scale only axis,
clip=false,
xmin=0,
xmax=100,
xlabel={Patch Retrieval mAP [\%]},
y dir=reverse,
ymin=0.6,
ymax=19.4,
ytick={1,2,3,4,5,6,7,8,9,10,11,12,13,14,15,16,17,18,19},
yticklabels={{\meanstd},{\resize},{\brief},{\orb},{\binboost},{\sift},{\rootsift},{\dcsiamts},{\dcsiam},{\tfratio},{\pdcsiamts},{\tfmargin},{\pdcsiam},{\deepdesc},{\ptfmargin},{\ptfratio},{\psift},{\prootsift},{\pdeepdesc}},
axis background/.style={fill=white},
xmajorgrids,
ymajorgrids
]
\addplot[xbar, bar width=0.8, fill=mycolor1, dashed, draw=black, area legend] table[row sep=crcr] {%
44.5531471758806	19\\
};
\addplot[forget plot, color=white!15!black, dashed] table[row sep=crcr] {%
0	0.6\\
0	19.4\\
};
\node[right, align=left]
at (axis cs:101,19) {44.55\%};
\addplot [color=mycolor2, draw=none, mark size=1.5pt, mark=*, mark options={solid, fill=black!60!mycolor2, mycolor2}, forget plot]
  table[row sep=crcr]{%
64.1412188601288	19\\
};
\addplot [color=mycolor3, draw=none, mark size=1.5pt, mark=*, mark options={solid, fill=black!60!mycolor3, mycolor3}, forget plot]
  table[row sep=crcr]{%
42.9242726687861	19\\
};
\addplot [color=mycolor4, draw=none, mark size=1.5pt, mark=*, mark options={solid, fill=black!60!mycolor4, mycolor4}, forget plot]
  table[row sep=crcr]{%
26.5939499987269	19\\
};
\addplot[xbar, bar width=0.8, fill=red!50!green, dashed, draw=black, area legend] table[row sep=crcr] {%
43.8416740642281	18\\
};
\addplot[forget plot, color=white!15!black, dashed] table[row sep=crcr] {%
0	0.6\\
0	19.4\\
};
\node[right, align=left]
at (axis cs:101,18) {43.84\%};
\addplot [color=mycolor2, draw=none, mark size=1.5pt, mark=*, mark options={solid, fill=black!60!mycolor2, mycolor2}, forget plot]
  table[row sep=crcr]{%
68.5693300831885	18\\
};
\addplot [color=mycolor3, draw=none, mark size=1.5pt, mark=*, mark options={solid, fill=black!60!mycolor3, mycolor3}, forget plot]
  table[row sep=crcr]{%
39.9622391919699	18\\
};
\addplot [color=mycolor4, draw=none, mark size=1.5pt, mark=*, mark options={solid, fill=black!60!mycolor4, mycolor4}, forget plot]
  table[row sep=crcr]{%
22.9934529175258	18\\
};
\addplot[xbar, bar width=0.8, fill=mycolor5, dashed, draw=black, area legend] table[row sep=crcr] {%
40.3551985831009	17\\
};
\addplot[forget plot, color=white!15!black, dashed] table[row sep=crcr] {%
0	0.6\\
0	19.4\\
};
\node[right, align=left]
at (axis cs:101,17) {40.36\%};
\addplot [color=mycolor2, draw=none, mark size=1.5pt, mark=*, mark options={solid, fill=black!60!mycolor2, mycolor2}, forget plot]
  table[row sep=crcr]{%
64.7150495230904	17\\
};
\addplot [color=mycolor3, draw=none, mark size=1.5pt, mark=*, mark options={solid, fill=black!60!mycolor3, mycolor3}, forget plot]
  table[row sep=crcr]{%
36.1159304916833	17\\
};
\addplot [color=mycolor4, draw=none, mark size=1.5pt, mark=*, mark options={solid, fill=black!60!mycolor4, mycolor4}, forget plot]
  table[row sep=crcr]{%
20.2346157345289	17\\
};
\addplot[xbar, bar width=0.8, fill=mycolor6, dashed, draw=black, area legend] table[row sep=crcr] {%
40.2319201598621	16\\
};
\addplot[forget plot, color=white!15!black, dashed] table[row sep=crcr] {%
0	0.6\\
0	19.4\\
};
\node[right, align=left]
at (axis cs:101,16) {40.23\%};
\addplot [color=mycolor2, draw=none, mark size=1.5pt, mark=*, mark options={solid, fill=black!60!mycolor2, mycolor2}, forget plot]
  table[row sep=crcr]{%
60.1796032233191	16\\
};
\addplot [color=mycolor3, draw=none, mark size=1.5pt, mark=*, mark options={solid, fill=black!60!mycolor3, mycolor3}, forget plot]
  table[row sep=crcr]{%
38.383551193587	16\\
};
\addplot [color=mycolor4, draw=none, mark size=1.5pt, mark=*, mark options={solid, fill=black!60!mycolor4, mycolor4}, forget plot]
  table[row sep=crcr]{%
22.1326060626801	16\\
};
\addplot[xbar, bar width=0.8, fill=mycolor7, dashed, draw=black, area legend] table[row sep=crcr] {%
40.0180214098068	15\\
};
\addplot[forget plot, color=white!15!black, dashed] table[row sep=crcr] {%
0	0.6\\
0	19.4\\
};
\node[right, align=left]
at (axis cs:101,15) {40.02\%};
\addplot [color=mycolor2, draw=none, mark size=1.5pt, mark=*, mark options={solid, fill=black!60!mycolor2, mycolor2}, forget plot]
  table[row sep=crcr]{%
60.4260729384789	15\\
};
\addplot [color=mycolor3, draw=none, mark size=1.5pt, mark=*, mark options={solid, fill=black!60!mycolor3, mycolor3}, forget plot]
  table[row sep=crcr]{%
37.8233891222823	15\\
};
\addplot [color=mycolor4, draw=none, mark size=1.5pt, mark=*, mark options={solid, fill=black!60!mycolor4, mycolor4}, forget plot]
  table[row sep=crcr]{%
21.8046021686592	15\\
};
\addplot[xbar, bar width=0.8, fill=mycolor1, draw=white!20!black, area legend] table[row sep=crcr] {%
39.8348077364854	14\\
};
\addplot[forget plot, color=white!15!black] table[row sep=crcr] {%
0	0.6\\
0	19.4\\
};
\node[right, align=left]
at (axis cs:101,14) {39.83\%};
\addplot [color=mycolor2, draw=none, mark size=1.5pt, mark=*, mark options={solid, fill=black!60!mycolor2, mycolor2}, forget plot]
  table[row sep=crcr]{%
56.8450447544166	14\\
};
\addplot [color=mycolor3, draw=none, mark size=1.5pt, mark=*, mark options={solid, fill=black!60!mycolor3, mycolor3}, forget plot]
  table[row sep=crcr]{%
38.5383747774446	14\\
};
\addplot [color=mycolor4, draw=none, mark size=1.5pt, mark=*, mark options={solid, fill=black!60!mycolor4, mycolor4}, forget plot]
  table[row sep=crcr]{%
24.121003677595	14\\
};
\addplot[xbar, bar width=0.8, fill=mycolor8, dashed, draw=black, area legend] table[row sep=crcr] {%
39.6773275783797	13\\
};
\addplot[forget plot, color=white!15!black, dashed] table[row sep=crcr] {%
0	0.6\\
0	19.4\\
};
\node[right, align=left]
at (axis cs:101,13) {39.68\%};
\addplot [color=mycolor2, draw=none, mark size=1.5pt, mark=*, mark options={solid, fill=black!60!mycolor2, mycolor2}, forget plot]
  table[row sep=crcr]{%
58.7973914196784	13\\
};
\addplot [color=mycolor3, draw=none, mark size=1.5pt, mark=*, mark options={solid, fill=black!60!mycolor3, mycolor3}, forget plot]
  table[row sep=crcr]{%
37.6726325292387	13\\
};
\addplot [color=mycolor4, draw=none, mark size=1.5pt, mark=*, mark options={solid, fill=black!60!mycolor4, mycolor4}, forget plot]
  table[row sep=crcr]{%
22.561958786222	13\\
};
\addplot[xbar, bar width=0.8, fill=mycolor7, draw=white!20!black, area legend] table[row sep=crcr] {%
39.3973202588247	12\\
};
\addplot[forget plot, color=white!15!black] table[row sep=crcr] {%
0	0.6\\
0	19.4\\
};
\node[right, align=left]
at (axis cs:101,12) {39.40\%};
\addplot [color=mycolor2, draw=none, mark size=1.5pt, mark=*, mark options={solid, fill=black!60!mycolor2, mycolor2}, forget plot]
  table[row sep=crcr]{%
59.197817844921	12\\
};
\addplot [color=mycolor3, draw=none, mark size=1.5pt, mark=*, mark options={solid, fill=black!60!mycolor3, mycolor3}, forget plot]
  table[row sep=crcr]{%
37.4682788633117	12\\
};
\addplot [color=mycolor4, draw=none, mark size=1.5pt, mark=*, mark options={solid, fill=black!60!mycolor4, mycolor4}, forget plot]
  table[row sep=crcr]{%
21.5258640682414	12\\
};
\addplot[xbar, bar width=0.8, fill=mycolor9, dashed, draw=black, area legend] table[row sep=crcr] {%
38.2343394266907	11\\
};
\addplot[forget plot, color=white!15!black, dashed] table[row sep=crcr] {%
0	0.6\\
0	19.4\\
};
\node[right, align=left]
at (axis cs:101,11) {38.23\%};
\addplot [color=mycolor2, draw=none, mark size=1.5pt, mark=*, mark options={solid, fill=black!60!mycolor2, mycolor2}, forget plot]
  table[row sep=crcr]{%
56.550619188255	11\\
};
\addplot [color=mycolor3, draw=none, mark size=1.5pt, mark=*, mark options={solid, fill=black!60!mycolor3, mycolor3}, forget plot]
  table[row sep=crcr]{%
36.7076182096344	11\\
};
\addplot [color=mycolor4, draw=none, mark size=1.5pt, mark=*, mark options={solid, fill=black!60!mycolor4, mycolor4}, forget plot]
  table[row sep=crcr]{%
21.4447808821826	11\\
};
\addplot[xbar, bar width=0.8, fill=mycolor6, draw=white!20!black, area legend] table[row sep=crcr] {%
37.6859862740109	10\\
};
\addplot[forget plot, color=white!15!black] table[row sep=crcr] {%
0	0.6\\
0	19.4\\
};
\node[right, align=left]
at (axis cs:101,10) {37.69\%};
\addplot [color=mycolor2, draw=none, mark size=1.5pt, mark=*, mark options={solid, fill=black!60!mycolor2, mycolor2}, forget plot]
  table[row sep=crcr]{%
56.6200894445833	10\\
};
\addplot [color=mycolor3, draw=none, mark size=1.5pt, mark=*, mark options={solid, fill=black!60!mycolor3, mycolor3}, forget plot]
  table[row sep=crcr]{%
35.8655668644687	10\\
};
\addplot [color=mycolor4, draw=none, mark size=1.5pt, mark=*, mark options={solid, fill=black!60!mycolor4, mycolor4}, forget plot]
  table[row sep=crcr]{%
20.5723025129807	10\\
};
\addplot[xbar, bar width=0.8, fill=mycolor8, draw=white!20!black, area legend] table[row sep=crcr] {%
34.8428311442408	9\\
};
\addplot[forget plot, color=white!15!black] table[row sep=crcr] {%
0	0.6\\
0	19.4\\
};
\node[right, align=left]
at (axis cs:101,9) {34.84\%};
\addplot [color=mycolor2, draw=none, mark size=1.5pt, mark=*, mark options={solid, fill=black!60!mycolor2, mycolor2}, forget plot]
  table[row sep=crcr]{%
53.8047467553443	9\\
};
\addplot [color=mycolor3, draw=none, mark size=1.5pt, mark=*, mark options={solid, fill=black!60!mycolor3, mycolor3}, forget plot]
  table[row sep=crcr]{%
32.2814485749621	9\\
};
\addplot [color=mycolor4, draw=none, mark size=1.5pt, mark=*, mark options={solid, fill=black!60!mycolor4, mycolor4}, forget plot]
  table[row sep=crcr]{%
18.442298102416	9\\
};
\addplot[xbar, bar width=0.8, fill=mycolor9, draw=white!20!black, area legend] table[row sep=crcr] {%
34.7640500172037	8\\
};
\addplot[forget plot, color=white!15!black] table[row sep=crcr] {%
0	0.6\\
0	19.4\\
};
\node[right, align=left]
at (axis cs:101,8) {34.76\%};
\addplot [color=mycolor2, draw=none, mark size=1.5pt, mark=*, mark options={solid, fill=black!60!mycolor2, mycolor2}, forget plot]
  table[row sep=crcr]{%
52.5469139535737	8\\
};
\addplot [color=mycolor3, draw=none, mark size=1.5pt, mark=*, mark options={solid, fill=black!60!mycolor3, mycolor3}, forget plot]
  table[row sep=crcr]{%
32.9199115651935	8\\
};
\addplot [color=mycolor4, draw=none, mark size=1.5pt, mark=*, mark options={solid, fill=black!60!mycolor4, mycolor4}, forget plot]
  table[row sep=crcr]{%
18.8253245328438	8\\
};
\addplot[xbar, bar width=0.8, fill=red!50!green, draw=white!20!black, area legend] table[row sep=crcr] {%
33.5575733935947	7\\
};
\addplot[forget plot, color=white!15!black] table[row sep=crcr] {%
0	0.6\\
0	19.4\\
};
\node[right, align=left]
at (axis cs:101,7) {33.56\%};
\addplot [color=mycolor2, draw=none, mark size=1.5pt, mark=*, mark options={solid, fill=black!60!mycolor2, mycolor2}, forget plot]
  table[row sep=crcr]{%
56.8381807641923	7\\
};
\addplot [color=mycolor3, draw=none, mark size=1.5pt, mark=*, mark options={solid, fill=black!60!mycolor3, mycolor3}, forget plot]
  table[row sep=crcr]{%
28.6813773879338	7\\
};
\addplot [color=mycolor4, draw=none, mark size=1.5pt, mark=*, mark options={solid, fill=black!60!mycolor4, mycolor4}, forget plot]
  table[row sep=crcr]{%
15.1531620286581	7\\
};
\addplot[xbar, bar width=0.8, fill=mycolor5, draw=white!20!black, area legend] table[row sep=crcr] {%
31.9790762627847	6\\
};
\addplot[forget plot, color=white!15!black] table[row sep=crcr] {%
0	0.6\\
0	19.4\\
};
\node[right, align=left]
at (axis cs:101,6) {31.98\%};
\addplot [color=mycolor2, draw=none, mark size=1.5pt, mark=*, mark options={solid, fill=black!60!mycolor2, mycolor2}, forget plot]
  table[row sep=crcr]{%
54.6865385030546	6\\
};
\addplot [color=mycolor3, draw=none, mark size=1.5pt, mark=*, mark options={solid, fill=black!60!mycolor3, mycolor3}, forget plot]
  table[row sep=crcr]{%
27.0338606194889	6\\
};
\addplot [color=mycolor4, draw=none, mark size=1.5pt, mark=*, mark options={solid, fill=black!60!mycolor4, mycolor4}, forget plot]
  table[row sep=crcr]{%
14.2168296658107	6\\
};
\addplot[xbar, bar width=0.8, fill=mycolor10, draw=white!20!black, area legend] table[row sep=crcr] {%
22.4546867509642	5\\
};
\addplot[forget plot, color=white!15!black] table[row sep=crcr] {%
0	0.6\\
0	19.4\\
};
\node[right, align=left]
at (axis cs:101,5) {22.45\%};
\addplot [color=mycolor2, draw=none, mark size=1.5pt, mark=*, mark options={solid, fill=black!60!mycolor2, mycolor2}, forget plot]
  table[row sep=crcr]{%
40.4117246468329	5\\
};
\addplot [color=mycolor3, draw=none, mark size=1.5pt, mark=*, mark options={solid, fill=black!60!mycolor3, mycolor3}, forget plot]
  table[row sep=crcr]{%
18.1046660704778	5\\
};
\addplot [color=mycolor4, draw=none, mark size=1.5pt, mark=*, mark options={solid, fill=black!60!mycolor4, mycolor4}, forget plot]
  table[row sep=crcr]{%
8.84766953558176	5\\
};
\addplot[xbar, bar width=0.8, fill=mycolor11, draw=white!20!black, area legend] table[row sep=crcr] {%
18.8464754804	4\\
};
\addplot[forget plot, color=white!15!black] table[row sep=crcr] {%
0	0.6\\
0	19.4\\
};
\node[right, align=left]
at (axis cs:101,4) {18.85\%};
\addplot [color=mycolor2, draw=none, mark size=1.5pt, mark=*, mark options={solid, fill=black!60!mycolor2, mycolor2}, forget plot]
  table[row sep=crcr]{%
36.8947770610578	4\\
};
\addplot [color=mycolor3, draw=none, mark size=1.5pt, mark=*, mark options={solid, fill=black!60!mycolor3, mycolor3}, forget plot]
  table[row sep=crcr]{%
13.549167323459	4\\
};
\addplot [color=mycolor4, draw=none, mark size=1.5pt, mark=*, mark options={solid, fill=black!60!mycolor4, mycolor4}, forget plot]
  table[row sep=crcr]{%
6.09548205668308	4\\
};
\addplot[xbar, bar width=0.8, fill=mycolor12, draw=white!20!black, area legend] table[row sep=crcr] {%
16.0268548034169	3\\
};
\addplot[forget plot, color=white!15!black] table[row sep=crcr] {%
0	0.6\\
0	19.4\\
};
\node[right, align=left]
at (axis cs:101,3) {16.03\%};
\addplot [color=mycolor2, draw=none, mark size=1.5pt, mark=*, mark options={solid, fill=black!60!mycolor2, mycolor2}, forget plot]
  table[row sep=crcr]{%
31.4816880615	3\\
};
\addplot [color=mycolor3, draw=none, mark size=1.5pt, mark=*, mark options={solid, fill=black!60!mycolor3, mycolor3}, forget plot]
  table[row sep=crcr]{%
11.5375438246943	3\\
};
\addplot [color=mycolor4, draw=none, mark size=1.5pt, mark=*, mark options={solid, fill=black!60!mycolor4, mycolor4}, forget plot]
  table[row sep=crcr]{%
5.06133252405634	3\\
};
\addplot[xbar, bar width=0.8, fill=mycolor13, draw=white!20!black, area legend] table[row sep=crcr] {%
13.1179822309423	2\\
};
\addplot[forget plot, color=white!15!black] table[row sep=crcr] {%
0	0.6\\
0	19.4\\
};
\node[right, align=left]
at (axis cs:101,2) {13.12\%};
\addplot [color=mycolor2, draw=none, mark size=1.5pt, mark=*, mark options={solid, fill=black!60!mycolor2, mycolor2}, forget plot]
  table[row sep=crcr]{%
26.0879928484614	2\\
};
\addplot [color=mycolor3, draw=none, mark size=1.5pt, mark=*, mark options={solid, fill=black!60!mycolor3, mycolor3}, forget plot]
  table[row sep=crcr]{%
9.09436209671969	2\\
};
\addplot [color=mycolor4, draw=none, mark size=1.5pt, mark=*, mark options={solid, fill=black!60!mycolor4, mycolor4}, forget plot]
  table[row sep=crcr]{%
4.17159174764579	2\\
};
\addplot[xbar, bar width=0.8, fill=lightgray, draw=white!20!black, area legend] table[row sep=crcr] {%
1.20074320957371	1\\
};
\addplot[forget plot, color=white!15!black] table[row sep=crcr] {%
0	0.6\\
0	19.4\\
};
\node[right, align=left]
at (axis cs:101,1) {1.20\%};
\addplot [color=mycolor2, draw=none, mark size=1.5pt, mark=*, mark options={solid, fill=black!60!mycolor2, mycolor2}, forget plot]
  table[row sep=crcr]{%
1.38685071490064	1\\
};
\addplot [color=mycolor3, draw=none, mark size=1.5pt, mark=*, mark options={solid, fill=black!60!mycolor3, mycolor3}, forget plot]
  table[row sep=crcr]{%
1.1912905951196	1\\
};
\addplot [color=mycolor4, draw=none, mark size=1.5pt, mark=*, mark options={solid, fill=black!60!mycolor4, mycolor4}, forget plot]
  table[row sep=crcr]{%
1.02408831870088	1\\
};
\end{axis}
\end{tikzpicture}%

%% file: conclusions.tex
\section{Conclusions}

With the advent of deep learning, the development of novel and more powerful local descriptors has accelerated tremendously. However, as we have shown in this paper, the benchmarks commonly used for evaluating such descriptors are inadequate, making comparisons unreliable. In the long run, this is likely to be detrimental to further research. In order to address this problem, we have introduced \dataset, a new public benchmark for local descriptors. The new benchmark is patch-based, removing many of the ambiguities that plagued the existing image-based benchmarks and favouring rigorous, reproducible, and large scale experimentation. This benchmark also improves on the limited data and task diversity present in other datasets, by considering many different scene and visual effects types, as well as three benchmark tasks close to practical applications of descriptors. 

Despite the multitask complexity of our benchmark suite, using the evaluation is easy as we provide open-source implementation of the protocols which can be used with minimal effort. \dataset can supersede datasets such as PhotoTourism and the older but still frequently used Oxford matching dataset, addressing their shortcomings and providing a valuable tool for researchers interested in local descriptors. \vspace{-1em}

\paragraph{Acknowledgements}
Karel Lenc is supported by ERC 677195-IDIU and Vassileios Balntas is
supported by FACER2VM EPSRC EP/N007743/1. We would like to thank
Giorgos Tolias for help with descriptor normalisation.

% \ifarxiv
% We would like to thank Giorgos Tolias for help with descriptor normalisation.
% \fi